\documentclass[letterpaper]{article} 

\usepackage{aaai2027,times}

\usepackage{amsmath,amsfonts,bm}

\def\eqref#1{equation~\ref{#1}}

\def\1{\bm{1}}

\DeclareMathAlphabet{\mathsfit}{\encodingdefault}{\sfdefault}{m}{sl}
\SetMathAlphabet{\mathsfit}{bold}{\encodingdefault}{\sfdefault}{bx}{n}

\usepackage[hyphens]{url}  
\usepackage{graphicx}  
\usepackage{wrapfig}
\usepackage{natbib}  
\usepackage{caption} 
\usepackage{algorithm}
\usepackage{algorithmic}

\usepackage{colortbl}

\usepackage[most]{tcolorbox}
 
\newtcolorbox{researchq}{
  enhanced,
  colback=black!4,           
  colframe=black!25,         
  boxrule=0.4pt,
  arc=3pt,                   
  left=6pt, right=6pt, top=5pt, bottom=5pt,
  boxsep=0pt,
  before skip=6pt, after skip=6pt,
}
\usepackage{bold-extra}
\usepackage{newfloat}
\usepackage{makecell}
\usepackage{listings}

\usepackage{booktabs}
\usepackage{multirow}
\usepackage{xcolor}
\definecolor{rebuttalblue}{HTML}{173F70}

\newif\ifshowrebuttal
\showrebuttalfalse
\DeclareRobustCommand{\rebuttal}[1]{%
  \ifshowrebuttal\textcolor{rebuttalblue}{#1}\else#1\fi}
\usepackage{amsmath,amssymb,amsthm}

\title{Recurrent Domain Incremental Adaptation of Time-Series Foundation Models}
\author{%
Sachith Abeywickrama\textsuperscript{1,2}
\quad
Peiliang Gong\textsuperscript{1}
\quad
Emadeldeen Eldele\textsuperscript{3}
\\[1mm]
\textbf{Min Wu\textsuperscript{2}}
\quad
\textbf{Chau Yuen\textsuperscript{1}}
\\[2mm]
\textsuperscript{1}Nanyang Technological University, Singapore
\\
\textsuperscript{2}Institute of Advanced Intelligence and Computing (IAIC), A*STAR, Singapore
\\
\textsuperscript{3}Department of Computer Science, Khalifa University, UAE
}

\iclrfinalcopy 
\begin{document}

\maketitle

\begin{abstract}
Pretrained Time-Series Foundation Models (TSFMs) provide strong representations for downstream classification tasks. Yet adapting a TSFM under changing operating conditions remains challenging. A model must improve on each newly observed condition without sacrificing conditions it has already learned, often from only a small labeled calibration set and without knowing the operating condition at inference time. We find that this difficulty has an important geometric structure. Under an operating-condition shift, the embedding space of a TSFM does not move uniformly: some local regions change substantially, while others remain nearly unchanged. Consequently, a global adaptation can unnecessarily move embeddings that were already correctly classified. To address this, we propose the novel {\textbf{PACT}} (\textbf{P}iecewise \textbf{A}daptation via \textbf{C}hart \textbf{T}ransport). PACT partitions the frozen source embedding space into local charts and learns lightweight residual transports for the regions affected by each new condition. An explicit identity path is always retained, and a learned gate softly selects among the identity and accumulated transports for each input. The pretrained TSFM encoder, classifier head, and previously learned transports remain fixed. With only 5\% labeled calibration data per arriving condition and no condition identity at test time, PACT maintains an average macro-F1 above 0.95 across a four-condition bearing-fault stream. Under homogeneous tasks and label spaces, similar retention-adaptation behavior is observed across public benchmarks. 
Code is available at \url{https://github.com/Sachithx/PACT}.
\end{abstract}

\section{Introduction}
\label{sec:intro}

\begin{figure}[h!]
    \centering
    \includegraphics[width=0.75\linewidth]{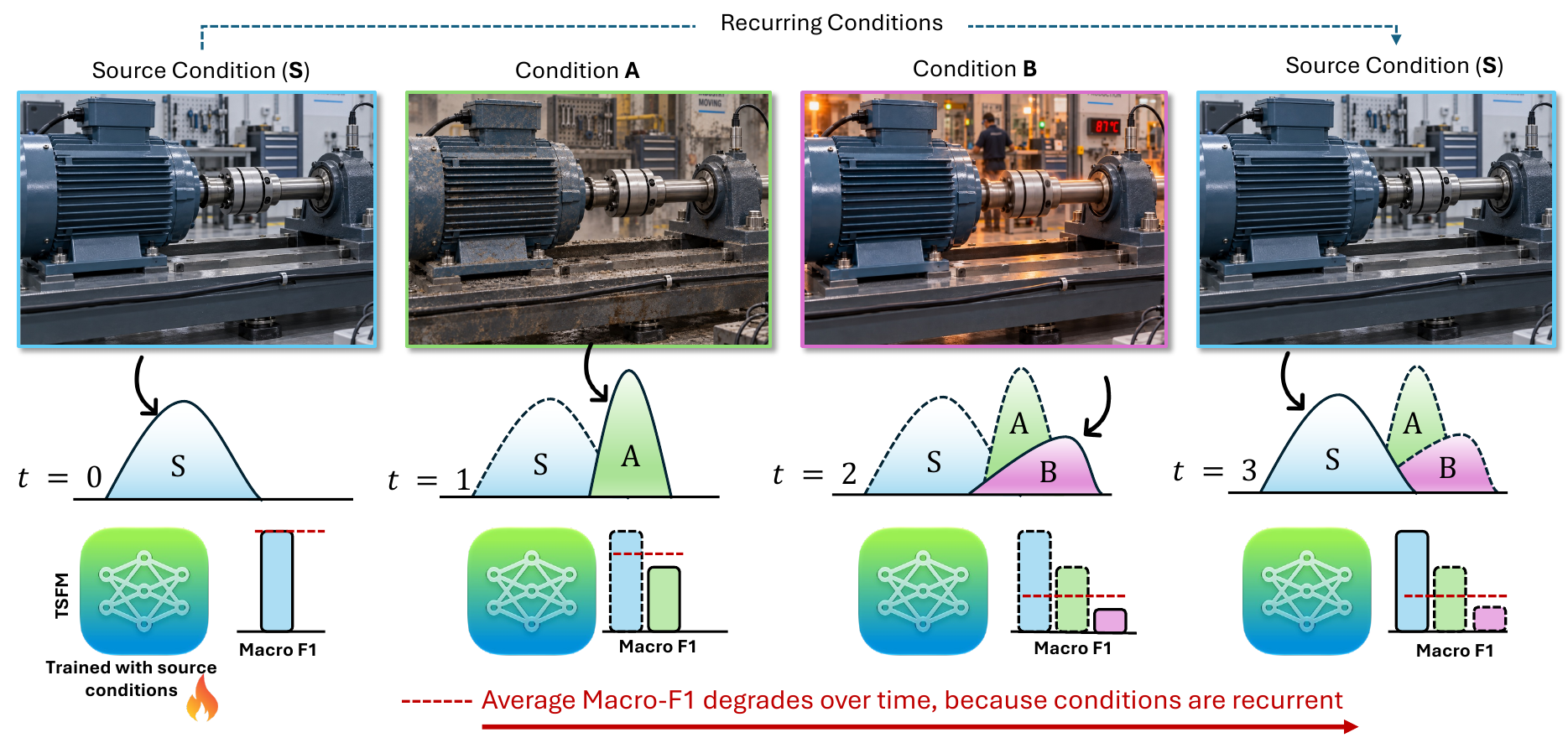}
    \caption{\textbf{Recurring operating conditions in machine fault diagnosis.} New conditions (A, B) arrive with few labels, while earlier ones, including the source S at t = 3, continue to recur at inference without condition identity. Average macro-F1 over seen conditions degrades in the absence of \textit{Recurrent Domain Incremental Adaptation.}
    }
    \label{fig:motivation}
    \vspace{-4pt}
\end{figure}

Pretrained time-series foundation models (TSFMs) have emerged as general-purpose representation learners for temporal data~\citep{goswami_moment_2024, moirai, ansari_chronos_2024, gong2026bridging, shi_time-moe_2025}. An effective, efficient way to leverage these models for downstream classification tasks is to freeze the pretrained TSFM encoder and train only a lightweight, task-specific head. This avoids retraining a large model and preserves the representation learned during pretraining. However, once initiated and deployed, the operating conditions may continue to change recurrently (Figure \ref{fig:motivation}).

Assuming a homogeneous task and label space, consider a fault-diagnosis system trained under a specific machine load or rotational speed \textit{(Source)}. During its service life, it may encounter new operating conditions. Similarly, when human subjects are closely matched in age and behavior, similar changes can arise across users in wearable sensing, subjects in physiological monitoring, or sensor configurations in mobile systems. These conditions often arise recurrently, with only a small amount of labeled data available for calibration. At inference time, the model may receive input samples from any condition encountered without being told which condition produced it. The model must therefore adapt to new conditions while still recognizing the old ones, given their recurrence.

\begin{wrapfigure}{r}{0.42\textwidth}
    \centering
    \includegraphics[width=\linewidth]{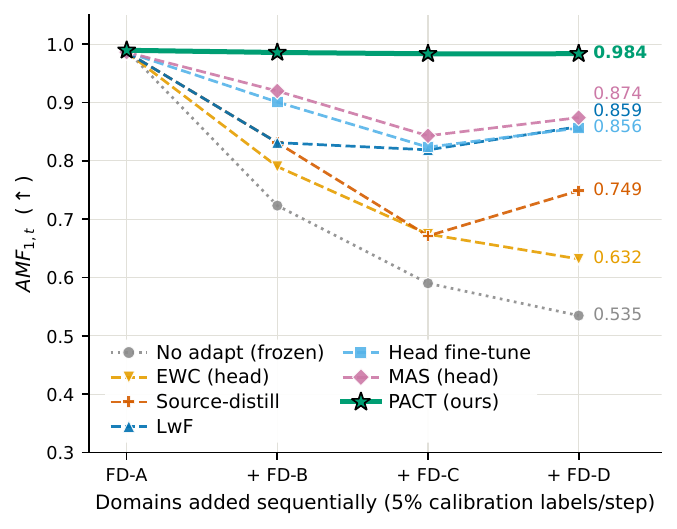}
    \caption{MOMENT-large on the fault-diagnosis benchmark.}
    \label{fig:bearing_example}
\end{wrapfigure}
This adaptation is challenging even when the pretrained TSFM representation is strong. Figure~\ref{fig:bearing_example} shows a representative bearing-fault stream. The classifier with MOMENT-Large TSFM backbone~\citep{goswami_moment_2024} performs well under its source condition, but its average Macro-F1 across all conditions drops sharply as conditions shift. Fine-tuning can recover performance on the new condition by updating model weights, but the same update changes parameters shared across all conditions and may degrade performance on earlier conditions. Regularization-based continual-learning methods reduce such changes, but this introduces the stability-plasticity trade-off: weak constraints allow adaptation but also lead to forgetting, whereas strong constraints preserve previous behavior at the expense of adaptation~\citep{ewc,mas,si,lwf}. Methods that allocate condition-specific parameters reduce direct interference, but commonly rely on hard routing, task cues, or reliable estimates of the condition identity~\citep{sprompts}. Furthermore, leveraging recent advances in strong TSFMs to address this core problem remains underexplored~\citep{tsfm_survey_ming,tsfm_survey_tpami}.

Existing approaches differ in how they control adaptation, but they leave a more fundamental question largely unanswered:

\begin{researchq}
\itshape ``Can a classifier built on a pretrained TSFM adapt to each new operating condition while retaining performance on the ones it already handles well?''
\end{researchq}

We study this question directly in the embedding space of frozen TSFMs. Our measurements reveal a simple but important phenomenon: domain condition shift is highly non-uniform across the representation space. Some local regions degrade substantially after a change in condition, whereas others remain nearly unaffected. Thus, the problem is not only how much to adapt, but also where adaptation is actually needed. Suppose an embedding already lies in a region that remains correctly represented under a new condition. For such an input, the desired correction may simply be the identity: nothing needs to change. An embedding in a strongly shifted region, in contrast, may require a non-zero correction before it is passed to the frozen classifier. A single domain-wide transformation changes both. But if distribution shift acts locally, this motivates testing whether chart-specific corrections improve on a single gated global transport.

To this end, we introduce PACT (Piecewise Adaptation via Chart Transport), a novel Domain Incremental Learning (DIL) method that leverages a pretrained TSFM. PACT partitions the source embedding space into local regions, which we call charts. When a new domain condition arrives, it learns a lightweight residual transport within each supported chart using the available labeled calibration examples. The adaptation then changes the embedding presented to the classifier head, not the representation model or its decision boundary. Once a condition is learned, its transports are frozen; later conditions add new transports without modifying the TSFM backbone, classifier head, or previously learned corrections. At inference time, the condition identity is unavailable, and an input may come from any condition. PACT therefore learns a chart-specific gate that softly combines the available condition transports. Importantly, this mixture always includes the unmodified source embedding as an explicit identity path. The gate can therefore apply a large correction when the evidence favors an adapted branch, combine several plausible corrections when conditions overlap, or remain close to the original representation when adaptation is unnecessary.

The identity path also makes the remaining source of degradation explicit. Because the TSFM encoder, classifier head, and previously learned transports never change, a newly arriving condition cannot overwrite these components. An earlier prediction can change only if routing moves its embedding through a non-identity transport. When the gate assigns all mass to the identity path, PACT exactly reproduces the frozen model's prediction. More generally, we derive a margin-based condition showing when a non-identity-routed correction is still small enough to preserve the predicted class. This is a conditional preservation result rather than a zero-forgetting guarantee: the remaining error comes from routing and the magnitude of the applied correction.

We evaluate PACT under recurrent condition shifts on five time-series domain adaptation benchmarks: bearing-fault diagnosis (FD), HHAR, Sleep-EDF, WISDM, and UCI-HAR. Under homogeneous labels and task spaces, with observed relative shifts from the source condition, the streams cover changes in operating condition, users, and subjects, and we never provide domain identity at inference time. We compare against frozen, fine-tuning, regularization, distillation, test-time adaptation, source-free adaptation, and domain-incremental baselines, all with matched calibration budgets. On the frozen Sundial backbone, PACT consistently outperforms the average macro-F1 on FD, WISDM, HHAR, and UCIHAR, while keeping forgetting below 0.05. The same mechanism transfers across eight classification-pretrained TSFMs, forecasting-pretrained TSFMs, and task-specific encoders, indicating that PACT is independent of the backbone architecture. Ablations further show that the two parts of the method are complementary: local transports provide adaptation, while learned identity-preserving routing is needed to avoid applying those corrections indiscriminately.

The contributions of this work are summarized as follows:

\begin{itemize}
    \item We show that condition shifts cause heterogeneous degradation across frozen embedding regions, motivating local correction.

    \item We propose a first-of-its-kind PACT that combines chart-specific residual transports with identity-aware routing for label-efficient domain-incremental adaptation, while keeping the encoder and head frozen.

    \item We evaluate PACT on five benchmarks and eight backbones, demonstrating effective adaptation and retention, and we include ablations that examine its key components.
\end{itemize}

\section{Related Work}
\label{sec:related}

Time-series foundation models provide transferable representations for downstream tasks with limited supervision \citep{tsfm_survey_ming,tsfm_survey_tpami}. Adapting these representations to changing conditions remains an active area of research. MAPU and its evidential extension, E-MAPU, adapt time-series feature extractors through temporal imputation and uncertainty-aware alignment \citep{mapu,emapu}. Alternatively, TATO adapts input transformations while keeping a forecasting TSFM frozen \citep{tato}. These approaches address target-domain adaptation; our focus is sequential classification with a fixed label set, where samples from earlier conditions remain relevant after each update. PACT therefore preserves both the pretrained encoder and source-trained head while learning corrections from small labeled calibration sets.

Regularization methods such as EWC, MAS, and SI constrain changes to important parameters, while Learning without Forgetting preserves previous outputs through distillation \citep{ewc,mas,si,lwf}. Test-time methods adapt using unlabeled observations: TENT minimizes prediction entropy, CoTTA addresses continually changing distributions, and EATA explicitly regularizes adaptation to reduce forgetting \citep{tent,cotta,eata}. For time-series classification, DualCD addresses domain-incremental learning through causal feature disentanglement and intervention \citep{dualcd}. These methods motivate the same adaptation-retention challenge, but differ from our adaptation constraints. PACT freezes previously learned corrections and updates their routing, preserving their parameters without guaranteeing unchanged predictions.

S-Prompts learns independent domain prompts and uses nearest-neighbor domain identification at inference \citep{sprompts}. RanPAC combines random projections with accumulated class statistics for class- and domain-incremental learning, while EASE expands representation subspaces for class-incremental learning \citep{ranpac,ease}. SEMA further learns mixtures of adapters with adaptive module expansion \citep{sema}. \rebuttal{PASS generates training features around stored class prototypes to preserve earlier classes during classifier updates \citep{pass}. FeCAM uses class-specific means and covariances to classify an observation by its distance to each class distribution \citep{fecam}. PACT assigns each embedding to a source-defined chart, where a learned gate blends the identity path with condition-specific corrections before the frozen head predicts the class (see Appendix~\ref{app:pact-prototypes} for the comparison with PASS and FeCAM).}

\section{Method}
\label{sec:method}

\subsection{Problem Formulation}
\label{sec:ProblemDefinition}

We consider recurrent domain-incremental time-series classification over a sequence of domain conditions $\mathcal{D}_0,\ldots,\mathcal{D}_T$ sharing a fixed label set. A pretrained encoder $f_\theta$ produces an aggregated, source-standardized embedding $z\in\mathbb{R}^{p}$ for each series, and a linear head $h_\phi$ is trained on the fully labeled source domain $\mathcal{D}_0$; both remain frozen during subsequent adaptation. Each new domain condition $\mathcal{D}_d$ provides a few-labeled calibration set comprising a fraction $\rho$ of its training split. Source and calibration embeddings are retained, while the raw series signal and class labels of the source or any previous domain condition are not stored or revisited. After observing the condition through step $t$, the objective is to classify samples from any domain condition without access to their identity, using an embedding-space correction $\Psi_t$: $\hat y=\arg\max_y[h_\phi(\Psi_t(z))]_y$. We assume a competent source representation and domain shifts that preserve the underlying labeling function and admit small residual corrections within adequately covered regions of the source embedding space. Furthermore, the concept shift is outside our scope.

\subsection{PACT: Piecewise Adaptation via Chart Transport}
\label{sec:pact_achitecture}

\begin{figure}[h!]
    \centering
    \includegraphics[width=0.75\linewidth]{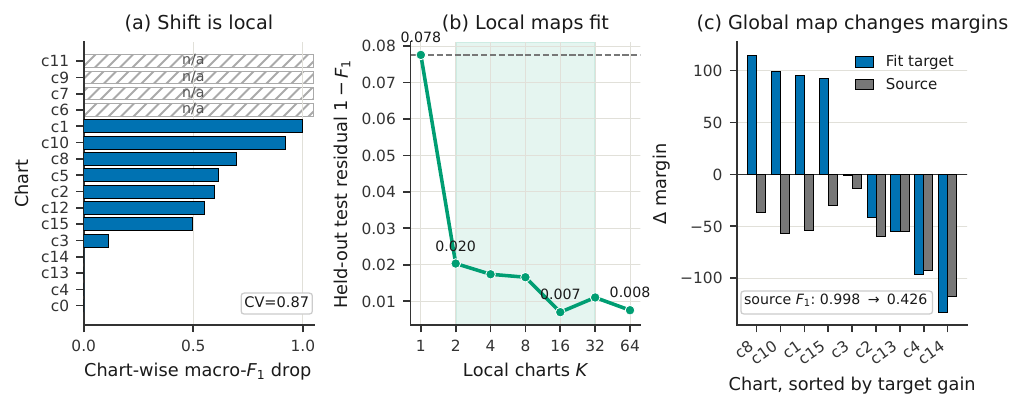}
    \caption{\rebuttal{Motivating local embedding corrections on FD with frozen TSLANet~\citep{eldele_tslanet_2024}, FD-A as source, and 5\% target calibration. We fit full-rank affine maps without routing. (a) Frozen-host degradation varies across source charts. (b) Held-out $1-F_1$ varies with chart count and model capacity. (c) A target-fitted global map improves some target margins while reducing source margins and source macro-$F_1$.}}  
    \label{fig:locality}
\end{figure}

\begin{figure*}[t]
    \centering
    \includegraphics[width=\textwidth]{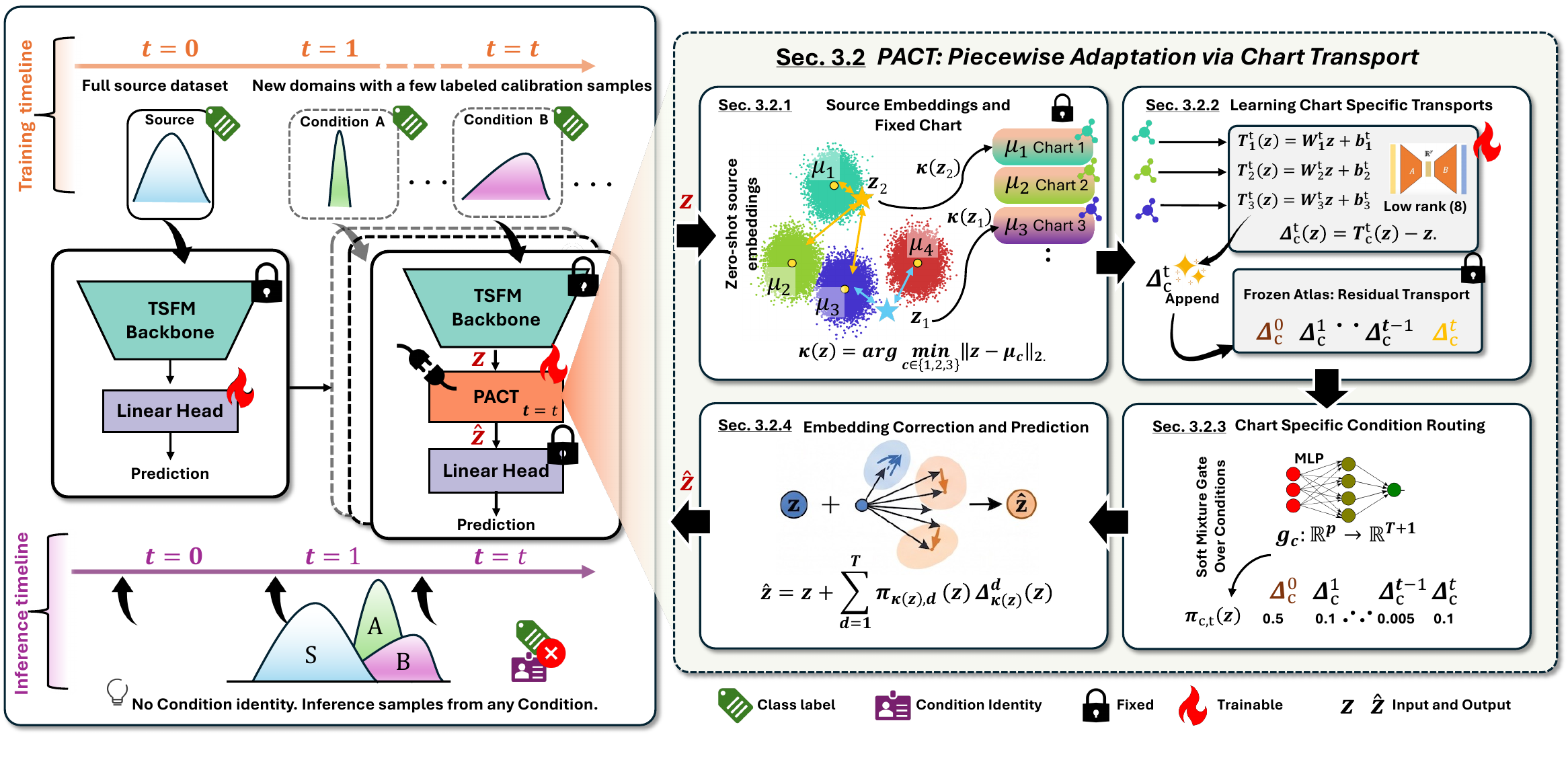}
    \caption{\textbf{Proposed PACT architecture.} Source embeddings define fixed nearest-centroid charts. Each arriving condition provides a small labeled calibration set to learn chart-specific residual transports, which are then frozen and appended to the transport bank. Chart-specific gates are refit using retained source and calibration embeddings. At inference, the selected chart's gate combines the identity branch and available transports, and the frozen head classifies the corrected embedding. The encoder and head remain unchanged throughout adaptation.
    }
    \label{fig:method}
\end{figure*}

\textsc{PACT} (Piecewise Adaptation via Chart Transport) adapts a frozen host through local residual corrections in embedding space. The observations in Figure~\ref{fig:locality} motivate this design: degradation varies across source-defined regions, local corrections reduce held-out error, and a target-fitted global correction can degrade source performance. Accordingly, \textsc{PACT} partitions the source embedding space into fixed charts, learns chart-specific transports for each arriving condition, and routes inputs through the accumulated transports without requiring condition identity. Figure~\ref{fig:method} summarizes the four stages.

\subsubsection{Source Embeddings and Fixed Charts}
\label{sec:charts}

Each time series is represented by an embedding $z\in\mathbb{R}^{p}$, obtained by averaging channel-wise window representations from the frozen encoder and standardizing them using source-fit statistics. We apply $k$-means to the source embeddings $Z_0$ to obtain centroids $\{\mu_c\}_{c=1}^{K}$. These centroids define Voronoi regions, termed \emph{charts}, with assignment
\begin{equation}
    \kappa(z)
    =
    \arg\min_{c\in\{1,\ldots,K\}}
    \lVert z-\mu_c\rVert_2^2.
    \label{eq:chart-assignment}
\end{equation}
The centroids and standardization statistics remain fixed across the domain condition stream. Each embedding is assigned to one chart, within which condition-specific corrections are learned and combined. Appendix Figure~\ref{fig:appendix-domain-structure} provides domain-colored projections and domain-centroid distances for the frozen TSLANet representations.

\subsubsection{Learning Chart-Specific Transports}
\label{sec:transports}

When domain condition $\mathcal{D}_d$ arrives, we learn a residual transport for each chart using its labeled calibration set $\mathcal{C}_d$. Our default parameterization is
\begin{equation}
    T_c^d(z)
    =
    z+\Delta_c^d(z),
    \qquad
    \Delta_c^d(z)
    =
    B_c^d A_c^d z,
    \label{eq:transport}
\end{equation}
where $A_c^d\in\mathbb{R}^{r\times p}$ and $B_c^d\in\mathbb{R}^{p\times r}$, with rank $r=8$ \citep{lora}. Each transport is initialized to the identity by setting its residual to zero. The source branch is always the fixed identity, $T_c^0(z)=z$.

During calibration, the new transport bank is trained directly, without routing:
\begin{equation}
    \mathcal{L}_{\mathrm{tr}}^{d}
    =
    \frac{1}{|\mathcal{C}_d|}
    \sum_{(x,y)\in\mathcal{C}_d}
    \ell_{\mathrm{CE}}
    \left(
        h_\phi\!\left(T_{\kappa(z)}^d(z)\right),y
    \right)
    +
    \lambda\sum_{c=1}^{K}
    \lVert B_c^d\rVert_F^2,
    \label{eq:transport-loss}
\end{equation}
where $z$ denotes the standardized embedding of $x$ and $\lambda=10^{-3}$. The regularizer penalizes the residual parameters but does not impose a pointwise bound on the embedding correction. Only the new transport parameters are optimized; the encoder, classifier head, and previously learned transports remain frozen. After training, we freeze the new bank and append it to the existing banks.

For the full-rank ablation, we instead use $T_c^d(z)=W_c^d z+b_c^d$, initialized with $W_c^d=I$ and $b_c^d=0$, and replace the regularizer with $\lambda\sum_c(\lVert W_c^d-I\rVert_F^2+\lVert b_c^d\rVert_2^2)$.

\subsubsection{Chart-Specific Domain Routing}
\label{sec:routing}

After observing domain conditions $\mathcal{D}_0,\ldots,\mathcal{D}_t$, each chart has a gate $g_c^{(t)}:\mathbb{R}^{p}\rightarrow\mathbb{R}^{t+1}$ that assigns weights to the source identity branch and the $t$ learned domain transports. Each gate is a two-layer MLP with hidden width $q=64$, ReLU activation, and dropout $0.2$. It is trained to predict domain identity from retained embeddings assigned to its chart, using cross-entropy and domain-balanced oversampling. Gates are refit after each condition arrival.

For an embedding assigned to chart $c$, the routing weights are
\begin{equation}
    \pi_{c,d}^{(t)}(z)
    =
    \left[
        \operatorname{softmax}
        \left(g_c^{(t)}(z)/\tau\right)
    \right]_d,
    \qquad d=0,\ldots,t,
    \label{eq:routing}
\end{equation}
with temperature $\tau=0.5$. A non-source branch with fewer than five calibration embeddings in a chart is masked before the softmax. The identity branch is always available. Thus, chart assignment is discrete, while routing among the available domain branches is soft.

\subsubsection{Embedding Correction and Prediction}
\label{sec:prediction}

At inference, \textsc{PACT} selects the nearest chart and combines its transport outputs according to the gate:
\begin{equation}
    \hat z
    =
    \sum_{d=0}^{t}
    \pi_{\kappa(z),d}^{(t)}(z)
    T_{\kappa(z)}^d(z)
    =
    z+
    \sum_{d=1}^{t}
    \pi_{\kappa(z),d}^{(t)}(z)
    \Delta_{\kappa(z)}^d(z).
    \label{eq:inference}
\end{equation}
The frozen classifier head then predicts $\hat y=\arg\max_y[h_\phi(\hat z)]_y$. Neither class labels nor condition identity are required. The identity branch allows the gate to retain the original embedding, while soft routing permits combinations of condition-specific corrections. Freezing earlier transports preserves their parameters, although predictions on earlier conditions can still change when the gates are refit. Appendix Figure~\ref{fig:appendix-transport} visualizes embedding corrections for two arriving HHAR domains.

\paragraph{Sequential updates and cost.}
Each new condition adds $2Kpr$ transport parameters and one output per gate. At step $t$, each gate contains $(p+1)q+(q+1)(t+1)$ parameters. Given an embedding, chart assignment costs $\mathcal{O}(Kp)$, gate evaluation costs $\mathcal{O}(pq+qt)$, and the transport mixture costs $\mathcal{O}(tpr)$. Gate refitting uses retained source and calibration embeddings, requiring $p(|Z_0|+\sum_{d=1}^{t}|Z_d|)$ floating-point values. We retain the source and calibration embeddings along with their domain indices for gate refitting; the raw time series and past class labels are discarded.

\section{Results}
\label{sec:experiments}

\subsection{Experimental Setup}
\label{sec:setup}

\paragraph{Datasets.}
We evaluate on five time-series classification benchmarks spanning distinct condition shift types: \textbf{FD} (Paderborn bearing-fault diagnosis; 4 operating conditions, 3 classes), \textbf{WISDM} (36 users, 6 activities), \textbf{HHAR} (9 users, 6 activities), \textbf{UCI-HAR} (30 subjects, 6 activities), and \textbf{Sleep-EDF} (20 subjects, 5 sleep stages) \citep{paderborn,kwapisz_wisdm_2011,stisen_hhar_2015,anguita_ucihar_2013,kemp_sleep_2000}.

\paragraph{Evaluation protocol.}
Each stream begins with a labeled source condition and introduces one target condition at a time. After each update, we evaluate all conditions seen so far on held-out examples without providing condition identity at inference. FD uses all four operating conditions, and HHAR uses all nine users. WISDM, UCI-HAR, and Sleep-EDF use nine subjects per stream, with subject composition and order varying across streams. Table~\ref{tab:main-results} reports the mean over three training seeds and four fixed streams per dataset. All methods use the same streams and data splits. Methods that use target class labels receive the same class-stratified calibration examples: 5\% of the arriving condition's training split on FD, HHAR, and Sleep-EDF, up to 450 examples on WISDM, and up to 100 on UCI-HAR. Appendix~\ref{app:pact-streams} lists the exact streams.

\paragraph{Backbones and PACT configuration.}
The main comparison uses frozen Sundial \citep{liu2025sundial}. We test backbone transfer with classification-pretrained Mantis and UniShape \citep{feofanov_mantis_2025,feofanov2026mantisv2,liu_unishape_2026}, forecasting-pretrained MOMENT, Chronos-Bolt, and Moirai \citep{goswami_moment_2024,ansari_chronos_2024,moirai}, and task-specific TSLANet and PatchTST \citep{eldele_tslanet_2024,nie_time_2022}. Each backbone is frozen during domain incremental adaptation. Condition identity is unavailable at inference. Unless stated otherwise, PACT uses rank-$8$ transports, gate temperature $\tau=0.5$, and the chart count $K$ chosen based on validation-split performance. Transports and gates are each trained for 200 epochs with the AdamW optimizer at a learning rate of $10^{-3}$; further implementation details are provided in the supplement (Appendix ~\ref{app:implementation}).

\paragraph{Baselines.}
All baselines operate on the same frozen Sundial embeddings and receive the same domain-condition orders, splits, and calibration-label budget. We compare against a non-adaptive frozen source model and nearest-class-mean (NCM) prototypes; test-time and source-free adaptation with CoTTA, EATA, MAPU, ACCUP, and TENT \citep{cotta,eata,mapu,accup,tent}; head fine-tuning and the regularization or distillation methods EWC, MAS, LwF, and source-distillation, which combines target cross-entropy with KL divergence from the frozen source head \citep{ewc,mas,lwf}; and the domain-incremental methods S-Prompts and DualCD \citep{sprompts,dualcd}. MAPU, ACCUP, and TENT are label-free; CoTTA, EATA, and the domain-incremental methods are adapted to this embedding/head setting, with details in the supplement. We also compare PACT with two rehearsal baselines. Embedding replay stores embeddings and their class labels. DER++ \citep{derpp} stores raw windows, class labels, and logits in a reservoir whose capacity is
set using the byte size of PACT's retained embedding array.

\paragraph{Metrics.}
Let $A_{t,j}$ denote macro-$F_1$ on the held-out test split of domain $j$ after adaptation through domain $t$. After the full stream, we report \emph{average macro-$F_1$}
$\mathrm{AMF_1}=\frac{1}{T+1}\sum_{j=0}^{T}A_{T,j}$ ($\uparrow$), together
with two lower-is-better forgetting measures,
\emph{relative forgetting}
$\mathrm{RF}=\frac{1}{T}\sum_{j=0}^{T-1}[p_j-A_{T,j}]_+/(p_j+\epsilon)$
and \emph{performance relative forgetting}
$\mathrm{PRF}=\frac{1}{T}\sum_{j=0}^{T-1}
(1-A_{T,j})[p_j-A_{T,j}]_+/(p_j+\epsilon)$, where
$p_j=\max_{j\le l<T}A_{l,j}$ is the peak score on domain $j$,
$[u]_+=\max(0,u)$, and $\epsilon=10^{-9}$. RF normalizes each peak-to-final drop by that peak, while PRF additionally weights the drop by how poor the final score is, following DualCD \citep{dualcd}.

\subsection{Main Results}
\label{sec:main-results}

\begin{table*}[h!]
\centering
\caption{
Performance comparison of PACT with a frozen Sundial backbone, averaged over the same four domain condition orders. 
Labeled methods receive the shared budget in Sec.~\ref{sec:setup}; \textsuperscript{\dag} denotes a label-free method. 
\textbf{Bold} marks the highest AMF$_1$ per dataset. \underline{Underline} marks the second highest AMF$_1$ per dataset.
Figure~\ref{fig:fig_rehearsal_comparison} compares PACT with embedding replay and DER++. 
}
\label{tab:main-results}
\small
\setlength{\tabcolsep}{3pt}
\renewcommand{\arraystretch}{1.2}

\resizebox{\textwidth}{!}{%
\begin{tabular}{c l ccc ccc ccc ccc ccc}
\toprule
& &
\multicolumn{3}{c}{FD} &
\multicolumn{3}{c}{WISDM} &
\multicolumn{3}{c}{HHAR} &
\multicolumn{3}{c}{UCIHAR} &
\multicolumn{3}{c}{Sleep-EDF} \\
\cmidrule(lr){3-5}
\cmidrule(lr){6-8}
\cmidrule(lr){9-11}
\cmidrule(lr){12-14}
\cmidrule(lr){15-17}
& \textbf{Method}
& \textbf{AMF$_1$}$\uparrow$ & \textbf{RF}$\downarrow$ & \textbf{PRF}$\downarrow$
& \textbf{AMF$_1$}$\uparrow$ & \textbf{RF}$\downarrow$ & \textbf{PRF}$\downarrow$
& \textbf{AMF$_1$}$\uparrow$ & \textbf{RF}$\downarrow$ & \textbf{PRF}$\downarrow$
& \textbf{AMF$_1$}$\uparrow$ & \textbf{RF}$\downarrow$ & \textbf{PRF}$\downarrow$
& \textbf{AMF$_1$}$\uparrow$ & \textbf{RF}$\downarrow$ & \textbf{PRF}$\downarrow$ \\
\midrule

\multirow{1}{*}{\textsc{No-DA}}
& Source-only
& 0.7084 & 0.0000 & 0.0000
& 0.5263 & 0.0000 & 0.0000
& 0.6768 & 0.0000 & 0.0000
& 0.5024 & 0.0000 & 0.0000
& 0.5313 & 0.0000 & 0.0000 \\
\cmidrule(lr){1-17}

\multirow{1}{*}{Few-shot}
& NCM prototypes
& 0.7336 & 0.0009 & 0.0001
& 0.6761 & 0.1245 & 0.0492
& 0.7145 & 0.0026 & 0.0007
& 0.5436 & 0.1099 & 0.0551
& 0.5043 & 0.0203 & 0.0099 \\
\cmidrule(lr){1-17}

\multirow{5}{*}{TTA/SFDA}
& CoTTA
& 0.7309 & 0.2625 & 0.1262
& 0.7793 & 0.1853 & 0.0703
& 0.8013 & 0.1123 & 0.0291
& 0.6704 & 0.2534 & 0.1041
& 0.5532 & 0.1413 & 0.0696 \\

& EATA
& 0.7546 & 0.2240 & 0.1021
& 0.7882 & 0.1739 & 0.0666
& 0.8063 & 0.1043 & 0.0267
& 0.6904 & 0.2265 & 0.0913
& 0.5693 & 0.1212 & 0.0580 \\

& MAPU\dag
& 0.5797 & 0.1555 & 0.0490
& 0.4081 & 0.0926 & 0.0541
& 0.5017 & 0.1707 & 0.0910
& 0.2555 & 0.1593 & 0.1184
& 0.3296 & 0.1527 & 0.1012 \\

& TENT\dag
& 0.6820 & 0.0540 & 0.0091
& 0.4121 & 0.0804 & 0.0446
& 0.5106 & 0.1509 & 0.0738
& 0.2470 & 0.1477 & 0.1094
& 0.3518 & 0.1318 & 0.0834 \\

& ACCUP\dag
& 0.6621 & 0.1132 & 0.0770 
& 0.3541 & 0.1734 & 0.1175 
& 0.5199 & 0.1680 & 0.0828 
& 0.3332 & 0.2010 & 0.1344 
& 0.5267 & 0.0166 & 0.0074 \\
\cmidrule(lr){1-17}

\multirow{7}{*}{Dom.-incr.}
& Head fine-tune
& 0.7709 & 0.2451 & 0.1061
& 0.7801 & 0.1844 & 0.0726
& 0.8072 & 0.1169 & 0.0297
& 0.6697 & 0.2571 & 0.1060
& 0.5724 & 0.1443 & 0.0679 \\

& EWC
& 0.7218 & 0.0267 & 0.0118
& 0.5290 & 0.0009 & 0.0005
& 0.7154 & 0.0043 & 0.0018
& 0.5121 & 0.0052 & 0.0023
& 0.5288 & 0.0028 & 0.0016 \\

& MAS
& 0.7848 & 0.1919 & 0.0924
& 0.7139 & 0.1702 & 0.0718
& 0.7706 & 0.1103 & 0.0425
& 0.6368 & 0.2341 & 0.1042
& 0.6114 & 0.0847 & 0.0382 \\

& LwF
& 0.7805 & 0.0799 & 0.0161
& \underline{0.7940} & 0.0882 & 0.0273
& 0.8217 & 0.0241 & 0.0059
& \underline{0.7351} & 0.0809 & 0.0275
& \textbf{0.6138} & 0.0488 & 0.0204 \\

& Source-distill
& 0.7296 & 0.1390 & 0.0739
& 0.6443 & 0.1782 & 0.0921
& 0.7361 & 0.0935 & 0.0370
& 0.6153 & 0.1455 & 0.0655
& 0.5761 & 0.0610 & 0.0288 \\

& S-Prompts
& 0.8227 & 0.0326 & 0.0067
& 0.6929 & 0.1300 & 0.0559
& 0.7151 & 0.0902 & 0.0305
& 0.5564 & 0.1093 & 0.0508
& 0.5196 & 0.0692 & 0.0350 \\

& DualCD
& \underline{0.8460} & 0.1771 & 0.0657
& 0.7810 & 0.1889 & 0.0640
& \underline{0.8296} & 0.0990 & 0.0257
& 0.7032 & 0.2360 & 0.0918
& 0.5796 & 0.1380 & 0.0643 \\
\cmidrule(lr){1-17}

\rowcolor{blue!10}

\multirow{1}{*}{Ours}
& \textbf{PACT}
& \textbf{0.9575} & 0.0021 & 0.0001
& \textbf{0.8774} & 0.0292 & 0.0085
& \textbf{0.8367} & 0.0210 & 0.0036
& \textbf{0.7864} & 0.0425 & 0.0101
& \underline{0.6129} & 0.0171 & 0.0100 \\
\bottomrule
\end{tabular}%
}
\end{table*}

\begin{figure}[h!]
    \centering
    \includegraphics[width=0.7\linewidth]{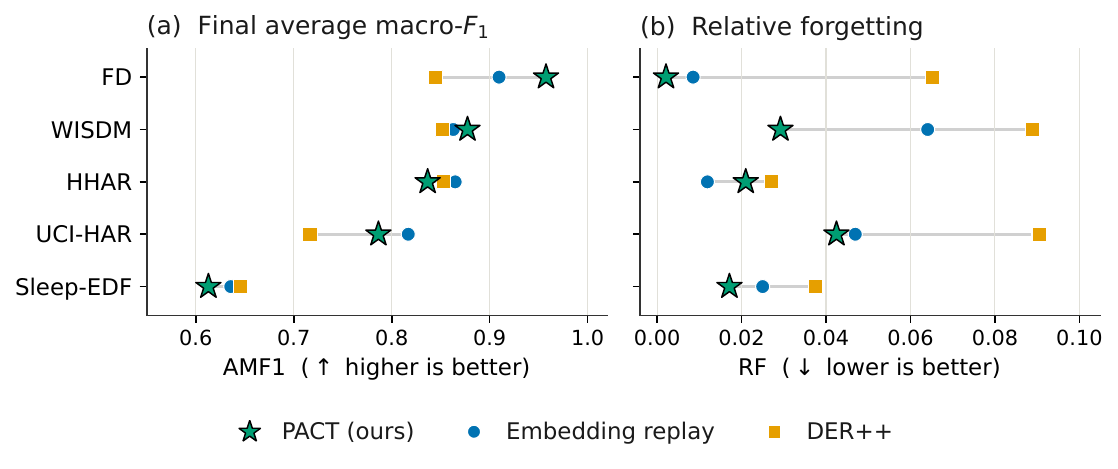}
    \caption{Comparison with rehearsal-based methods. 
    PACT has lower RF than both replay methods on FD, WISDM, UCI-HAR, and Sleep-EDF; embedding replay has lower RF on HHAR
    }
    \label{fig:fig_rehearsal_comparison}
\end{figure}

\paragraph{Main comparison.}
Among the methods in Table~\ref{tab:main-results}, PACT achieved the highest final AMF$_1$ on FD, WISDM, HHAR, and UCI-HAR. Its largest gains were on FD and WISDM, where it exceeded the strongest baseline shown by 11.15 and 8.34 percentage points, respectively. The margin was smaller on HHAR (0.71 points), and LwF led PACT by 0.09 points on Sleep-EDF. PACT's relative forgetting ranged from 0.0021 on FD to 0.0425 on UCI-HAR. \rebuttal{Figure~\ref{fig:fig_rehearsal_comparison} compares PACT with two rehearsal-based methods on the same four streams per dataset. PACT achieved higher final AMF1 than both replay methods on FD and WISDM, exceeding embedding replay by 4.79 and 1.45 percentage points, respectively. Embedding replay led PACT by 2.81 points on HHAR and 3.05 points on UCI-HAR, while DER++ led by 3.24 points on Sleep-EDF. PACT achieved these results while discarding past class labels after each transport bank was fitted. When both methods were limited to 512 historical embeddings,
PACT retained a 4.38-point AMF$_1$ advantage over embedding replay on
FD, while replay led on the other four datasets
(Table~\ref{tab:pact-history}).}

\paragraph{Source retention and condition revisits.} Across four streams per dataset, PACT had a smaller final decline in source macro-$F_1$ than embedding replay and head fine-tuning on all five datasets. The decline ranged from 0.02 percentage points on FD to 6.16 points on UCI-HAR (Appendix~\ref{app:pact-retention}). \rebuttal{In a separate simulated FD revisit, held-out macro-$F_1$ on condition A remained at 0.9748 after learning B and C, and macro-$F_1$ on condition B remained at 0.9137 after learning D. Both revisits used the learned gate without condition identity at inference. The benchmark streams evaluate recurrence by testing held-out examples from previously seen conditions after each update.}

\begin{wrapfigure}{r}{0.5\textwidth}
    \centering
    \includegraphics[width=1\linewidth]{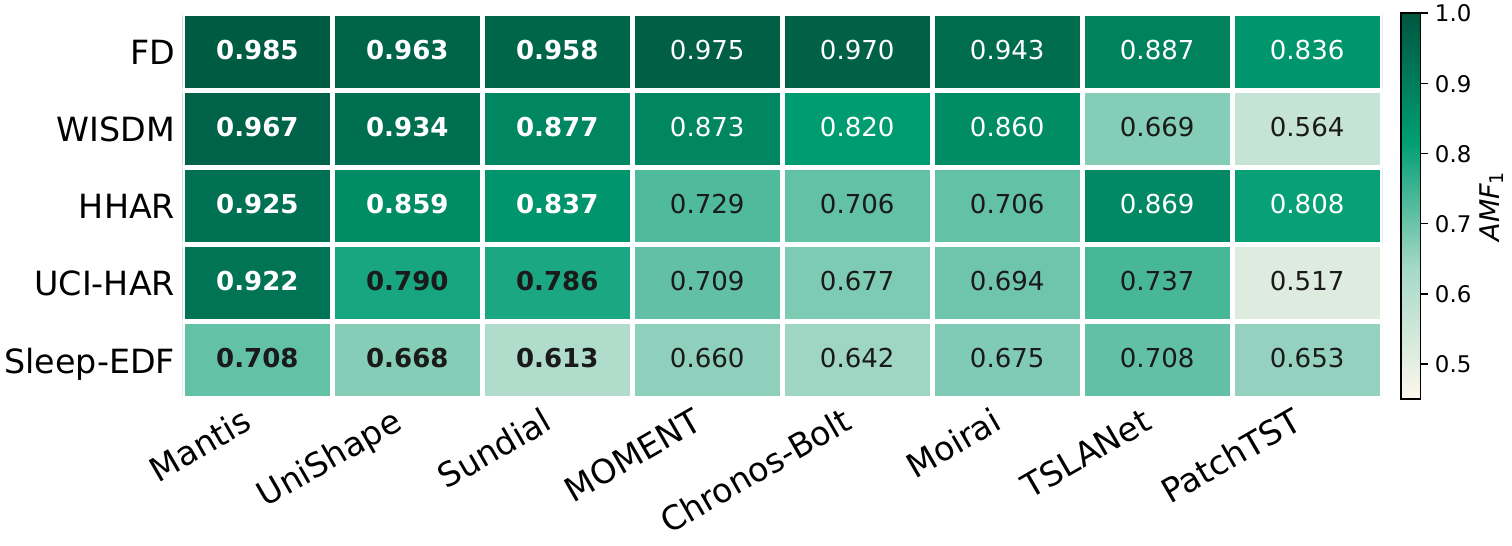}
    \caption{\textbf{PACT across eight backbones.} Each cell is the mean AMF$_1$. The columns cover classification-pretrained, forecasting-pretrained, and task-specific encoders.}
    \label{fig:backbones}
\end{wrapfigure}
\paragraph{Transfer across backbones.} Because PACT requires only an embedding and a frozen head, it applies unchanged to task-specific encoders and to forecasting-pretrained TSFMs used purely as feature extractors. Figure~\ref{fig:backbones} reports this across eight hosts. Mantis is strongest on FD, WISDM, HHAR, and UCI-HAR, reaching AMF$_1$ of 0.985, 0.967, 0.925, and 0.922, while TSLANet is also strong on Sleep-EDF at 0.708. The spread within the dataset is nevertheless substantial: UCI-HAR ranges from 0.517 (PatchTST) to 0.922 (Mantis). PACT is therefore architecturally portable, but the separability of the frozen representation bounds what any embedding-space correction can recover. Appendix Figure~\ref{fig:appendix-retention} shows per-domain performance matrices across eight backbones for selected domain orders.

\subsection{Ablation Study}
\label{sec:ablation}

\begin{figure}[h!]
    \centering
    \includegraphics[width=0.75\linewidth]{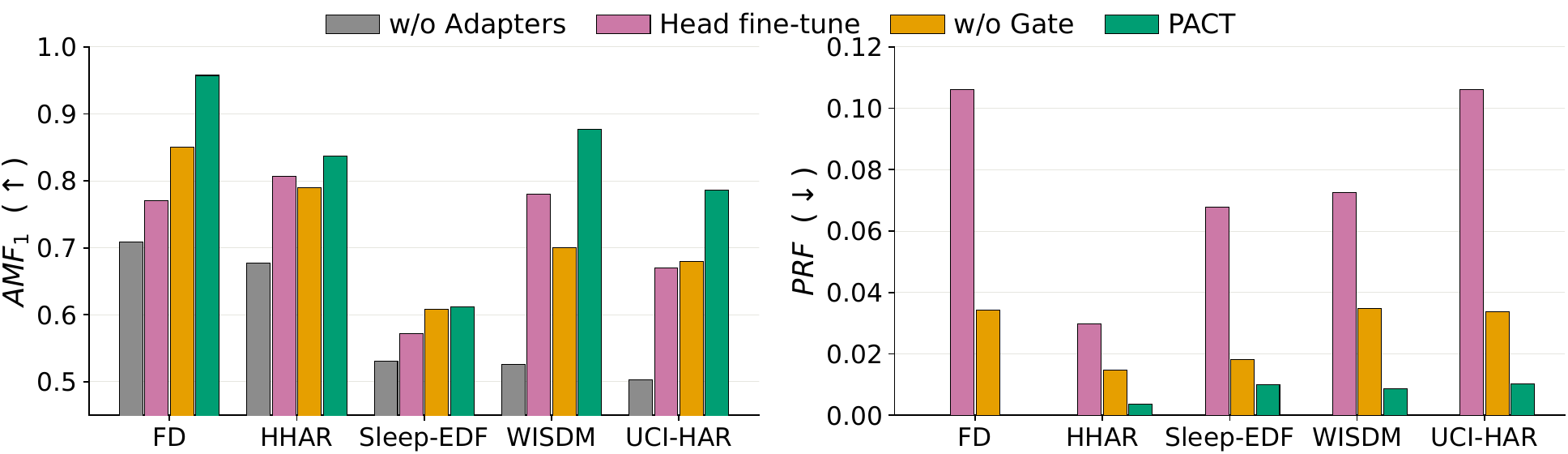}
    \caption{\textbf{Component ablation on frozen Sundial.} Removing transports gives the non-adaptive source model; head fine-tuning updates only the classifier; removing the gate uses an ungated transport mixture. Full PACT combines rank-8 chart transports with learned soft routing. 
    }
    \label{fig:components}
\end{figure}

\paragraph{Impact of residual transports and routing.} 
Figure~\ref{fig:components} separates plasticity from retention. Head fine-tuning increases AMF$_1$ relative to the frozen model but incurs the highest PRF across all datasets, adapting at the cost of retention. Ungated transports raise AMF$_1$ further, yet still forget by correcting indiscriminately. Adding the learned gate gives both the best AMF$_1$ and the lowest PRF among adaptive variants on all five datasets, so local correction and identity-preserving routing are complementary rather than alternatives. The effect is largest on FD and WISDM and smallest for Sleep-EDF, consistent with their more overlapping embedding distributions.

\paragraph{Effect of chart count and placement.} \rebuttal{With rank-8 transports, the identity branch, and learned routing retained throughout, increasing the chart count from $K=1$ to $K=4$ improved AMF$_1$ on FD and all three activity datasets (Table~\ref{tab:chart-locality}). FD improved further at $K=8$, while the three activity datasets achieved their highest AMF$_1$ at $K=4$. At fixed transport and gate capacity, source $k$-means outperformed random-projection partitions on all four streams for each of the five tested dataset--$K$ configurations, including a second UCI-HAR comparison at $K=8$ (Appendix~\ref{app:pact-placement}).}

\begin{table}[th]
\centering
\caption{\rebuttal{Effect of chart count and placement on frozen Sundial.
Entries report final AMF$_1$ averaged over four streams.
The chart-count sweep uses source $k$-means throughout.
The placement columns report a separate paired experiment with
matched chart counts and identical transport and gate capacity.
Bold marks the higher or highest AMF$_1$ within each comparison.}}
\label{tab:chart-locality}
\scriptsize
\begin{tabular}{@{}l ccc c cc@{}}
\toprule
& \multicolumn{3}{c}{Chart-count sweep}
& \multicolumn{3}{c}{Equal-capacity placement} \\
\cmidrule(lr){2-4}\cmidrule(l){5-7}
\textbf{Dataset}
& $K=1$ & $K=4$ & $K=8$
& $K$ & Source $k$-means & Random projection \\
\midrule
FD
& 0.9386 & 0.9484 & \textbf{0.9575}
& 8 & \textbf{0.9575} & 0.8936 \\
WISDM
& 0.8717 & \textbf{0.8774} & 0.8420
& 4 & \textbf{0.8774} & 0.8078 \\
HHAR
& 0.8318 & \textbf{0.8369} & 0.8090
& 4 & \textbf{0.8367} & 0.7839 \\
UCI-HAR
& 0.7743 & \textbf{0.7862} & 0.7725
& 4 & \textbf{0.7864} & 0.7553 \\
\bottomrule
\end{tabular}
\end{table}

\begin{wrapfigure}{r}{0.5\textwidth}
    \centering
    \includegraphics[width=1\linewidth]{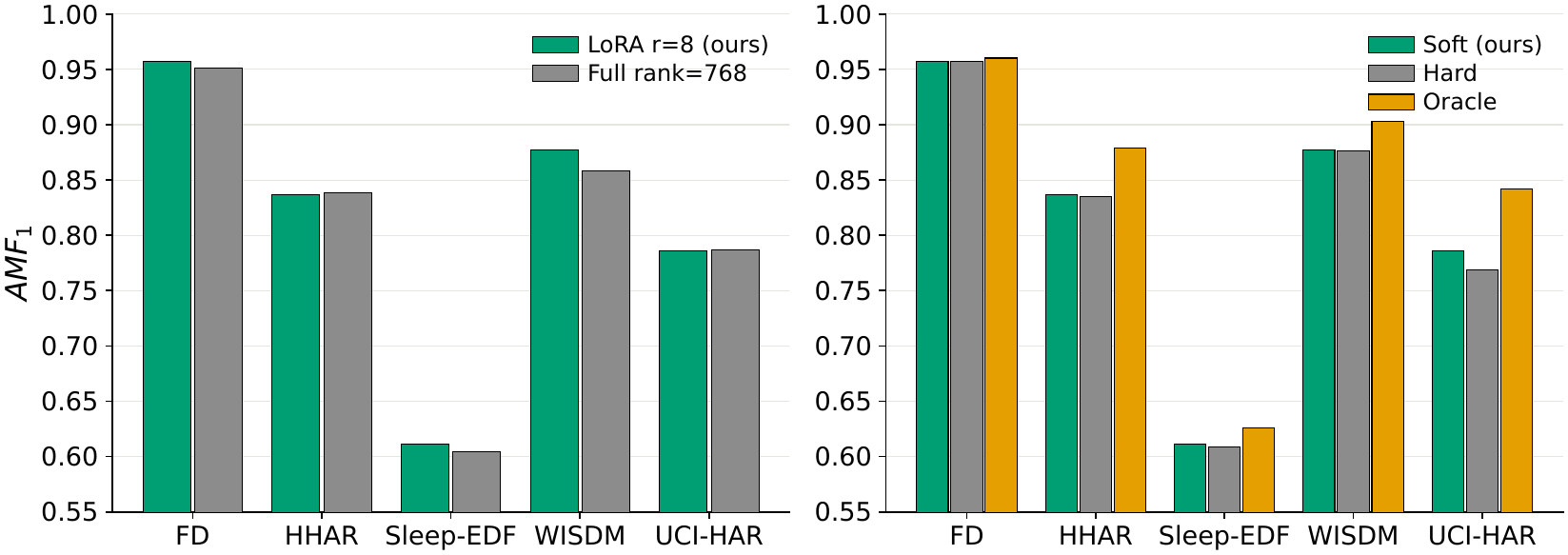}
    \caption{\textbf{Transport rank and routing.} Left: rank-8 residual transports versus full-rank (768). Right: learned soft routing, hard argmax routing, and oracle routing with the true condition identity. Bars report mean AMF$_1$ over four orders.}
    \label{fig:rank-routing}
\end{wrapfigure}
\paragraph{Impact of transport rank and routing strategy.} Figure~\ref{fig:rank-routing} compares rank-8 and full-rank transports and evaluates three routing strategies. Rank-8 transports achieved comparable or higher AMF$_1$ across the five datasets, using $2Kpr$ trainable transport parameters per condition compared with $Kp(p+1)$ for full-rank affine maps. The frozen classifier distinguishes at most six classes, so its predictions depend on at most five independent logit differences. \rebuttal{The rank comparison therefore establishes parameter efficiency for this classification task. Learned soft routing matched or exceeded hard routing in the reported comparisons. The larger gaps to oracle routing on HHAR and UCI-HAR indicate greater scope for improvement in condition assignment on these datasets.}

\subsection{Label Efficiency and Calibration Robustness}
\label{sec:robustness}

\begin{figure}
    \centering
    \includegraphics[width=0.6\linewidth]{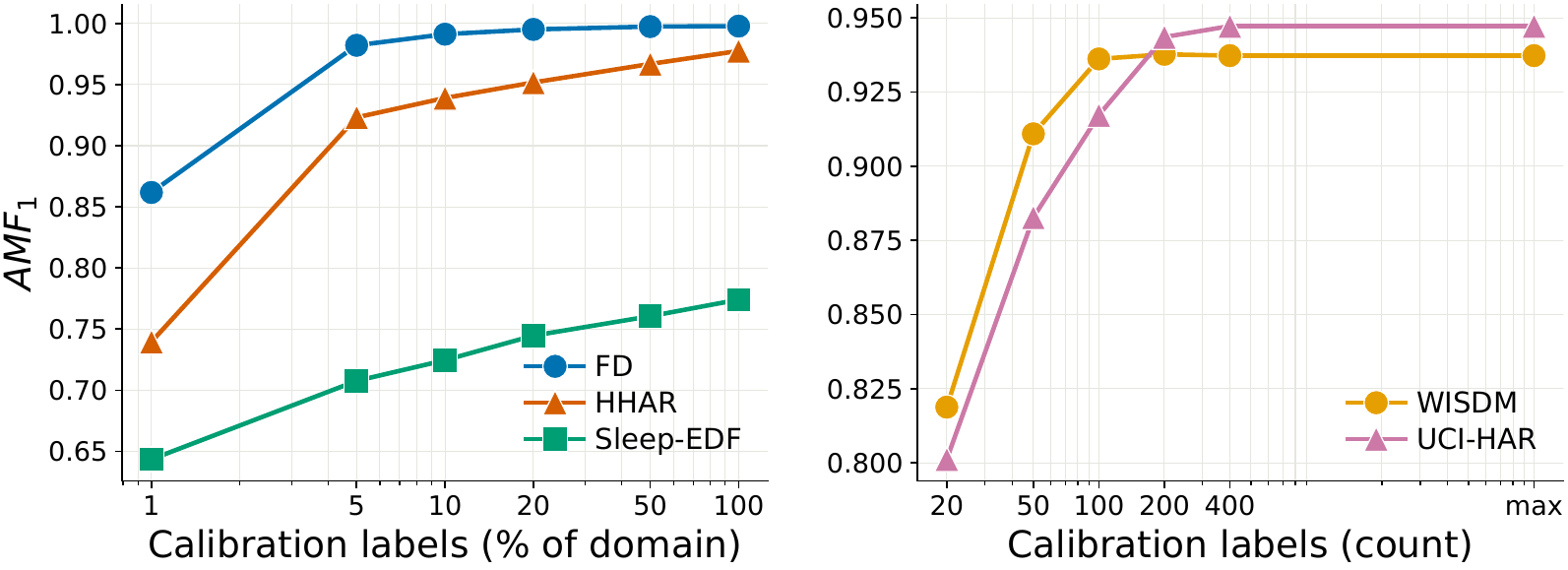}
    \caption{\textbf{Label efficiency with frozen Mantis.} Left: calibration as a percentage on FD, HHAR, and Sleep-EDF. Right: a fixed number of labels on WISDM and UCI-HAR.}
    \label{fig:label-efficiency}
\end{figure}
On frozen Mantis, Figure~\ref{fig:label-efficiency} shows steep gains at small budgets. With 5\% labels, AMF$_1$ reaches 0.982 on FD and 0.923 on HHAR; Sleep-EDF improves more gradually from 0.643 at 1\% to 0.774 with all labels. For fixed-count streams, WISDM reaches 0.936 with 100 labels and changes little beyond 200 labels, whereas UCI-HAR continues to improve through roughly 400 labels. The curves support the few-label setting and show that the useful budget depends on the difficulty of the domain condition. We also varied the class composition of the calibration set on frozen Sundial while holding its size fixed. Assigning 85\% of the labels to one class reduced AMF$_1$ by 9.08 points on FD and 1.38 points on UCI-HAR. When class~0 was omitted, its recall was 0.995 on FD and 0.154 on UCI-HAR across four streams (Appendix~\ref{app:pact-imbalance}).

\subsection{Conditional Prediction Guarantees}
\label{sec:guarantees}

Fix step $t\geq1$ and chart $c=\kappa(z)$, and abbreviate $\pi_j=\pi_{c,j}^{(t)}(z)$. For $h_\phi(u)=Hu+a$, choose $L_h>0$ with $L_h\geq\max_y\|H_{y,:}\|_2$, and define $m_y(u)=[h_\phi(u)]_y-\max_{k\neq y}[h_\phi(u)]_k$.

\paragraph{Proposition 1 (Prediction preservation).}
Let $y_0$ be the unique frozen-host prediction and
$B_t(z)=\max_{1\leq j\leq t}\|\Delta_c^j(z)\|_2$. Then
\begin{equation}
    \|\hat z-z\|_2\leq(1-\pi_0)B_t(z).
\end{equation}
The prediction remains $y_0$ whenever
$(1-\pi_0)B_t(z)<m_{y_0}(z)/(2L_h)$.
Thus, correctness is preserved if the original prediction is correct;
$\pi_0=1$ recovers the frozen embedding exactly.

\paragraph{Proposition 2 (Target correctness).}
Let $z$ belong to domain $1\leq d\leq t$ with true label $y$,
and let $s$ be a same-class source embedding in chart $c$ with $m_y(s)>0$. Define
\begin{equation}
\begin{aligned}
    e_d&=\|T_c^d(z)-s\|_2,\\
    R_{d,t}&=\max_{\substack{0\leq j\leq t\\j\neq d}}
    \|T_c^j(z)-T_c^d(z)\|_2.
\end{aligned}
\end{equation}
Then $\|\hat z-s\|_2\leq e_d+(1-\pi_d)R_{d,t}$, and PACT
predicts $y$ whenever
\begin{equation}
    e_d+(1-\pi_d)R_{d,t}<\frac{m_y(s)}{2L_h}.
\end{equation}
Oracle routing removes the branch-mixture term. These are sufficient conditions after fitting, not guarantees of successful learning or zero forgetting across updates. Their failure does not imply misclassification. Proofs appear in Appendix~\ref{app:guarantees}. \rebuttal{We measured how often Proposition~1 certifies an initially correct source prediction. Coverage was 99.8\% on FD, 97.5\% on HHAR, and 71.1\% on UCI-HAR. All certified predictions were preserved. The FD diagnostic uses one stream, and the HHAR and UCI-HAR diagnostics use four streams each (Appendix~\ref{app:pact-certificates}).}

\section{Conclusion}
\label{sec:conclusion}

We introduced PACT, a frozen-host method for recurrent domain incremental adaptation of TSFMs. Across five benchmarks, the frozen-Sundial comparison yields a strong AMF$_1$ for PACT while keeping forgetting below 0.05. The eight-backbone study shows that the mechanism transfers across architectures and highlights the importance of representation quality: Mantis leads on all datasets. Component, rank, and routing ablations support the three design choices of local transport, low rank, and soft identity-preserving routing.

\section{Limitations and Future Work}

PACT relies on the quality of the frozen representation and the coverage of its source-defined charts. If the encoder fails to distinguish classes under a new condition, embedding corrections have limited information with which to recover them. Target examples far from source support may also share a chart despite requiring different corrections. Adaptive chart refinement and soft chart assignment could improve coverage and reduce sensitivity near chart boundaries.

Retention also depends on routing. Overlapping conditions remain difficult to distinguish, as reflected in the oracle-routing gaps on HHAR and UCI-HAR. Although the identity branch preserves access to the original classifier, gate refitting can change earlier predictions. Improved routing and calibration could help preserve source predictions while maintaining adaptation to new conditions.

PACT retains source and calibration embeddings with condition assignments, and its transport bank grows as conditions arrive. The bounded-memory experiments show that reducing the retained history can affect performance, particularly on the activity datasets. More effective embedding selection and transport consolidation are therefore useful directions. Finally, calibration imbalance and missing classes can reduce adaptation quality. Future work could investigate label-efficient calibration strategies and evaluation on naturally recurring streams with longer deployment horizons.







\bibliography{aaai2027}
\bibliographystyle{aaai2027}

\appendix
\section{Proofs of Conditional Prediction Guarantees}
\label{app:guarantees}

Each logit is $L_h$-Lipschitz, implying
$m_y(v)\geq m_y(u)-2L_h\|v-u\|_2$.
For Proposition~1, the triangle inequality gives
\[
\|\hat z-z\|_2
\leq\sum_{j=1}^{t}\pi_j\|\Delta_c^j(z)\|_2
\leq(1-\pi_0)B_t(z).
\]
For Proposition~2, use
\[
\hat z-s=T_c^d(z)-s+
\sum_{j\neq d}\pi_j\big(T_c^j(z)-T_c^d(z)\big).
\]
The triangle inequality yields the stated distance bound.
Applying the margin inequality establishes both prediction conditions.
Verifying the source anchor's label and correctness requires labeled
analysis data; these labels are not required at inference.

\section{Implementation and Reproducibility Details}
\label{app:implementation}

\subsection{Domain streams and calibration protocol}
\label{app:pact-streams}

\rebuttal{Each stream begins with a labeled source condition. After adapting to each arriving condition, we evaluate all conditions observed so far on held-out examples. FD uses all four operating conditions, and HHAR uses all nine users. For WISDM, UCI-HAR, and Sleep-EDF, each stream contains nine subjects. Both subject composition and order can vary across the four streams for these datasets. We fixed the streams across methods and training seeds. Table~\ref{tab:pact-stream-ids} gives the exact sequences.}

\begin{table}[h!]
\centering
\caption{Domain identifiers in arrival order. The first identifier in
each sequence denotes the labeled source.}
\label{tab:pact-stream-ids}
\small
\setlength{\tabcolsep}{5pt}
\renewcommand{\arraystretch}{1.08}
\begin{tabular}{@{}lcl@{}}
\toprule
Dataset & Stream & Domain sequence \\
\midrule
FD
& 1 & $(0,2,3,1)$ \\
& 2 & $(2,0,1,3)$ \\
& 3 & $(2,1,0,3)$ \\
& 4 & $(2,1,3,0)$ \\
\addlinespace
WISDM
& 1 & $(10,24,20,0,3,1,22,14,2)$ \\
& 2 & $(12,23,7,9,5,15,31,28,10)$ \\
& 3 & $(24,11,0,23,13,21,16,35,3)$ \\
& 4 & $(30,3,0,31,20,24,19,13,17)$ \\
\addlinespace
HHAR
& 1 & $(1,6,4,7,8,5,0,2,3)$ \\
& 2 & $(5,1,6,3,2,7,4,0,8)$ \\
& 3 & $(5,8,0,3,2,6,4,1,7)$ \\
& 4 & $(6,0,8,2,5,1,3,4,7)$ \\
\addlinespace
UCI-HAR
& 1 & $(14,24,1,27,26,10,12,7,9)$ \\
& 2 & $(19,14,6,12,4,29,9,18,21)$ \\
& 3 & $(24,14,19,17,6,5,7,3,30)$ \\
& 4 & $(6,20,18,30,12,24,25,3,17)$ \\
\addlinespace
Sleep-EDF
& 1 & $(0,5,18,9,3,10,6,1,14)$ \\
& 2 & $(1,5,9,0,4,3,18,8,13)$ \\
& 3 & $(15,6,17,5,8,9,18,19,2)$ \\
& 4 & $(18,5,12,7,10,17,0,3,1)$ \\
\bottomrule
\end{tabular}
\end{table}

\rebuttal{Calibration examples are sampled by class from each arriving condition's training split. All methods that use target class labels receive the same examples. The budget is 5\% on FD, HHAR, and Sleep-EDF, up to 450 examples on WISDM, and up to 100 on UCI-HAR. The fixed-count caps accommodate differences in the number of available examples per subject. Condition identity is unavailable at inference.}

\subsection{Transport and gate optimization}
\label{app:pact-fitting}

The main comparison uses frozen Sundial embeddings. We average representations over evenly spaced, full-resolution windows and channels, then standardize them using statistics fitted on source training embeddings. Source $k$-means centroids define the charts and remain fixed throughout the stream. The dataset-specific configurations use $K=8$ on FD, $K=4$ on WISDM, HHAR, and UCI-HAR, and $K=1$ on Sleep-EDF. Experiments using a common chart count report $K$ separately.

Each arriving condition adds a bank of rank-8 transports initialized with zero residuals. We fit the new bank for 200 epochs using AdamW with a learning rate of $10^{-3}$. The encoder, source classifier, and previously learned transports remain frozen. Equation~(3) penalizes $\sum_c \lVert B_c^d\rVert_F^2$ with coefficient $10^{-3}$. The initial implementation used $\sum_c \lVert B_c^d A_c^d\rVert_F^2$. At the dataset-specific chart counts, rerunning the experiment with the penalty in Eq.~(3) reproduced the archived AMF$_1$ scores to within 0.0013.

Each chart gate is a two-layer MLP with 64 hidden units, ReLU activation, and dropout 0.2. After each arrival, we refit the gates on retained embeddings using condition identifiers, cross-entropy, and domain-balanced oversampling. Gate fitting uses AdamW for 200 epochs with a learning rate of $10^{-3}$. At inference, the softmax temperature is $\tau=0.5$ by default. A target branch is masked when its condition contributes fewer than five calibration embeddings to the chart. The identity branch is always available.

\subsection{Baseline implementations and retained information}
\label{app:pact-baseline-details}

\rebuttal{All methods use the same frozen Sundial representations. MAPU, ACCUP, and TENT use no target class labels. The S-Prompts-inspired variant fits a linear head for each condition and selects the head whose condition mean is nearest to the input embedding. The DualCD-inspired variant learns a linear feature mask and classifier with intra- and inter-domain losses. The EATA-inspired variant updates the head using entropy minimization and a Fisher anchor. The CoTTA-inspired variant uses perturbed embeddings and an exponential-moving-average teacher. Supervised baseline head fits use the shared calibration examples for 80 epochs. Baseline hyperparameters were tuned on the corresponding validation splits.}

\rebuttal{PACT retains source and calibration embeddings with condition assignments for gate refitting. It discards past class labels after fitting the corresponding transport bank. Embedding replay retains embeddings and class labels. DER++ retains raw time-series windows, class labels, and logits in a reservoir. We sized the DER++ reservoir against the final historical embedding-array byte budget for each stream. PACT's centroids, transport parameters, gates, and metadata contribute additional persistent storage. Each arriving condition adds $2Kpr$ transport parameters for embedding dimension $p$ and residual rank $r$.}

\paragraph{Computation Resources.}
All experiments were conducted on a Linux machine equipped with two NVIDIA RTX 6000 Ada Generation GPUs, each with 48\,GB of GPU memory, using NVIDIA driver version 570.124.06.

\section{Additional Results}

\subsection{Variation across streams and training seeds}
\label{app:pact-variability}

\rebuttal{We report two sources of variation. To measure stream variability, we fix the training seed and compute the standard deviation of final AMF$_1$ across the four streams. To measure seed variability, we first average over the four streams within each seed, then compute the mean and standard deviation across those seed-level averages. On WISDM, UCI-HAR, and Sleep-EDF, stream variability reflects differences in both subject composition and order.}

\begin{table}[h!]
\centering
\caption{Variation in final AMF$_1$ on frozen Sundial. Stream SD is the standard deviation of PACT across four streams. The final two columns report mean $\pm$ standard deviation over paired training seeds. The comparator is embedding replay on FD, WISDM, HHAR, and UCI-HAR, and DER++ on Sleep-EDF.}
\label{tab:pact-variability}
\small
\setlength{\tabcolsep}{5pt}
\renewcommand{\arraystretch}{1.12}
\begin{tabular}{@{}lcccc@{}}
\toprule
Dataset & Stream SD & Seeds & PACT & Comparator \\
\midrule
FD        & 0.0106 & 3 & $0.9575\pm0.0022$ & $0.9041\pm0.0061$ \\
WISDM     & 0.0197 & 2 & $0.8775\pm0.0173$ & $0.8621\pm0.0008$ \\
HHAR      & 0.0214 & 3 & $0.8367\pm0.0022$ & $0.8591\pm0.0044$ \\
UCI-HAR   & 0.0144 & 2 & $0.7910\pm0.0046$ & $0.8145\pm0.0024$ \\
Sleep-EDF & 0.0316 & 3 & $0.6129\pm0.0023$ & $0.6482\pm0.0023$ \\
\bottomrule
\end{tabular}
\end{table}

\rebuttal{On FD, PACT exceeded embedding replay by 5.34 AMF$_1$ points across three seeds. The WISDM ordering changed when averaged over two seeds, with replay ahead by 1.54 points. Embedding replay also achieved higher mean AMF$_1$ on HHAR and UCI-HAR, and DER++ led on Sleep-EDF.}

\subsection{Replay under matched historical embedding budgets}
\label{app:pact-replay}

\rebuttal{{We compare PACT and embedding replay under two storage budgets. In the full-history experiment, both methods retain embeddings from the same source and calibration examples. Under the bounded-history condition, each method retains at most 512 historical embeddings. Table~\ref{tab:pact-history} reports the paired reruns and the corresponding full-history capacities.}}

\begin{table}[h!]
\centering
\caption{Historical storage and final AMF$_1$ in the matched-history reruns. Scores average four streams. DER++ slots count raw windows with their labels and logits. The 512-embedding cap applies to PACT and embedding replay.}
\label{tab:pact-history}
\small
\setlength{\tabcolsep}{4pt}
\renewcommand{\arraystretch}{1.12}
\begin{tabular}{@{}l cc cc cc@{}}
\toprule
& \multicolumn{2}{c}{Full-history capacity}
& \multicolumn{2}{c}{Full history}
& \multicolumn{2}{c}{512 embeddings} \\
\cmidrule(lr){2-3}\cmidrule(lr){4-5}\cmidrule(l){6-7}
Dataset & Embeddings & DER++ slots & PACT & Replay & PACT & Replay \\
\midrule
FD & 2,104 & 315
   & 0.9575 & 0.9096 & 0.9556 & 0.9118 \\
WISDM & 978--1,280 & 1,916--2,507
      & 0.8774 & 0.8629 & 0.8206 & 0.8461 \\
HHAR & 1,630--1,708 & 3,193--3,346
     & 0.8367 & 0.8648 & 0.8252 & 0.8622 \\
UCI-HAR & $\approx$1,026--1,066 & 679--705
        & 0.7864 & 0.8169 & 0.7513 & 0.8078 \\
Sleep-EDF & 1,927--2,524 & 492--644
          & 0.6129 & 0.6357 & 0.6003 & 0.6255 \\
\bottomrule
\end{tabular}
\end{table}

\rebuttal{On FD, PACT led embedding replay by 4.79 AMF$_1$ points with full history and by 4.38 points under the 512-embedding cap. Under the cap, PACT's FD relative forgetting was also 0.0158 lower. WISDM showed greater sensitivity to eviction. PACT led by 1.45 points with full history, while replay led by 2.55 points under the cap. Replay achieved higher AMF$_1$ on HHAR, UCI-HAR, and Sleep-EDF under both budgets.}

\subsection{Chart placement at fixed capacity}
\label{app:pact-placement}

\rebuttal{To test whether chart placement contributes beyond the number of trainable parameters, we compare source $k$-means with a fixed random projection whose source quantiles define $K$ bins. Within each pair, the two methods use the same chart count, transport rank, gate capacity, identity branch, and four streams.}

\begin{table}[h!]
\centering
\caption{Equal-capacity partition comparison. Scores are final AMF$_1$
averaged over four paired streams. The last column counts streams in
which source $k$-means achieved higher AMF$_1$.}
\label{tab:pact-placement}
\small
\setlength{\tabcolsep}{5pt}
\renewcommand{\arraystretch}{1.12}
\begin{tabular}{@{}lcccc@{}}
\toprule
Dataset & $K$ & Source $k$-means & Random projection & Streams won \\
\midrule
FD       & 8 & 0.9575 & 0.8936 & 4/4 \\
WISDM    & 4 & 0.8774 & 0.8078 & 4/4 \\
HHAR     & 4 & 0.8367 & 0.7839 & 4/4 \\
UCI-HAR  & 4 & 0.7864 & 0.7553 & 4/4 \\
UCI-HAR  & 8 & 0.7725 & 0.7023 & 4/4 \\
\bottomrule
\end{tabular}
\end{table}

\rebuttal{Across the five configurations, the mean advantage of source $k$-means ranged from 3.11 to 7.02 AMF$_1$ points. The two UCI-HAR configurations reuse the same four streams. At $K=4$ on UCI-HAR, every target cell in the random partition met the minimum-support threshold. Source $k$-means still achieved a 3.11-point gain.}

\begin{table}[h!]
\centering
\caption{Relative forgetting in the chart-count sweep. Scores average four streams. All configurations retain rank-8 transports, learned soft routing, and the identity branch.}
\label{tab:pact-chart-rf}
\small
\renewcommand{\arraystretch}{1.12}
\begin{tabular}{@{}lccc@{}}
\toprule
Dataset & $K=1$ & $K=4$ & $K=8$ \\
\midrule
FD      & 0.0018 & 0.0021 & 0.0022 \\
WISDM   & 0.0317 & 0.0292 & 0.0458 \\
HHAR    & 0.0193 & 0.0171 & 0.0142 \\
UCI-HAR & 0.0549 & 0.0436 & 0.0462 \\
\bottomrule
\end{tabular}
\end{table}

Chart count affected adaptation and forgetting differently. On HHAR, increasing $K$ from four to eight reduced RF from 0.0171 to 0.0142, while final AMF$_1$ fell from 0.8369 to 0.8090. On FD, increasing $K$ from one to eight raised final AMF$_1$ from 0.9386 to 0.9575, with RF remaining between 0.0018 and 0.0022.

\subsection{Comparison with prototype-based methods}
\label{app:pact-prototypes}

We compare PACT with a frozen-feature adaptation of PASS and a pooled FeCAM classifier using the same Sundial embeddings, calibration budgets, four streams, and three seeds. The PASS variant augments training features around stored class prototypes. Freezing the encoder disables feature distillation, and we omit the original image-rotation objective. FeCAM pools class statistics across observed conditions and uses covariance shrinkage and normalization. PACT uses a common $K=4$ for this comparison.

\begin{table}[h!]
\centering
\caption{Mean final AMF$_1$ for the fixed-$K$ prototype comparison. The table gives datasets with reported scores for all three methods.}
\label{tab:pact-prototypes}
\small
\setlength{\tabcolsep}{6pt}
\renewcommand{\arraystretch}{1.12}
\begin{tabular}{@{}lccc@{}}
\toprule
Dataset & PACT ($K=4$) & PASS adaptation & Pooled FeCAM \\
\midrule
FD      & 0.9507 & 0.8392 & 0.9195 \\
WISDM   & 0.8576 & 0.8084 & 0.8297 \\
UCI-HAR & 0.7941 & 0.7629 & 0.7939 \\
\bottomrule
\end{tabular}
\end{table}

PACT exceeded pooled FeCAM by 3.12 points on FD and 2.79 points on WISDM. Their means differed by 0.02 points on UCI-HAR. On HHAR, pooled FeCAM led PACT by 3.99 points.

\subsection{Source retention and condition revisits}
\label{app:pact-retention}

\rebuttal{Let $A_{t,0}$ denote macro-$F_1$ on held-out source examples after the $t$th target update. We report the change in source performance as $100(A_{t,0}-A_{0,0})$ percentage points, averaged over four streams. This quantity tracks the source throughout adaptation. RF summarizes peak-to-final losses across previously seen conditions.}

\begin{table}[h!]
\centering
\caption{Initial source macro-$F_1$ and its change after each PACT
update, in percentage points. FD has three target updates.}
\label{tab:pact-trajectory}
\small
\setlength{\tabcolsep}{3pt}
\renewcommand{\arraystretch}{1.12}
\begin{tabular}{@{}l r rrrrrrrr@{}}
\toprule
& & \multicolumn{8}{c}{Target update} \\
\cmidrule(l){3-10}
Dataset & Initial & 1 & 2 & 3 & 4 & 5 & 6 & 7 & 8 \\
\midrule
FD & .9876
& $-.06$ & $-.02$ & $-.02$ & -- & -- & -- & -- & -- \\
WISDM & .9345
& $+.03$ & $-.39$ & $-.50$ & $-.69$ & $-.52$ & $-.72$ & $-1.12$ & $-1.40$ \\
HHAR & .9673
& $-.13$ & $-.20$ & $-.05$ & $-.09$ & $-.09$ & $.00$ & $-.24$ & $-.10$ \\
UCI-HAR & .9357
& $-2.90$ & $-3.61$ & $-4.71$ & $-6.68$ & $-5.27$ & $-5.92$ & $-4.92$ & $-6.16$ \\
Sleep-EDF & .7591
& $+.11$ & $-.24$ & $-.10$ & $-.24$ & $-.43$ & $-.30$ & $-.53$ & $-.14$ \\
\bottomrule
\end{tabular}
\end{table}

\begin{table}[h!]
\centering
\caption{Final change in source macro-$F_1$ from its initial value,
in percentage points, averaged over four streams.}
\label{tab:pact-endpoint}
\small
\setlength{\tabcolsep}{6pt}
\renewcommand{\arraystretch}{1.12}
\begin{tabular}{@{}lrrr@{}}
\toprule
Dataset & PACT & Embedding replay & Head fine-tuning \\
\midrule
FD        & $-0.02$ & $-0.33$  & $-31.76$ \\
WISDM     & $-1.40$ & $-10.22$ & $-34.70$ \\
HHAR      & $-0.10$ & $-1.13$  & $-10.38$ \\
UCI-HAR   & $-6.16$ & $-9.64$  & $-38.09$ \\
Sleep-EDF & $-0.14$ & $-4.39$  & $-15.26$ \\
\bottomrule
\end{tabular}
\end{table}

\rebuttal{PACT had a smaller final source decline than both comparators on all five datasets. UCI-HAR showed the largest decline for PACT. Source macro-$F_1$ fell 6.68 points below its starting value after the fourth update and ended 6.16 points below it. On FD, HHAR, and Sleep-EDF, mean source macro-$F_1$ stayed within 0.53 points of its initial value at every update.}

\rebuttal{The public subject streams evaluate retention by retesting earlier conditions after each update. We also simulated two explicit revisits on FD. Held-out A retained macro-$F_1$ of 0.9748 after adaptation to B and C, and held-out B retained 0.9137 after adaptation to D. Both revisits used the learned gate without a condition identifier.}

\subsection{Identity routing and source prediction preservation}
\label{app:pact-certificates}

\rebuttal{We evaluate Proposition~1 on source examples that the frozen classifier predicted correctly. Certificate coverage is the fraction of those examples satisfying the sufficient preservation condition. We also measure the fraction of source examples assigned identity weight $\pi_0>0.5$ and the fraction of target chart--condition cells masked at the final update.}

\begin{table}[h!]
\centering
\caption{Routing and preservation diagnostics (\%). FD uses the
$C\to B\to A\to D$ stream at $K=8$. HHAR and UCI-HAR use all four
streams at $K=4$.}
\label{tab:pact-certificates}
\small
\setlength{\tabcolsep}{5pt}
\renewcommand{\arraystretch}{1.12}
\begin{tabular}{@{}lrrr@{}}
\toprule
Diagnostic & FD & HHAR & UCI-HAR \\
\midrule
Source examples with $\pi_0>0.5$ & 99.9 & 97.5 & 77.9 \\
Originally correct source examples certified & 99.8 & 97.5 & 71.1 \\
Certified predictions preserved & 100.0 & 100.0 & 100.0 \\
Originally correct source predictions preserved & 99.9 & 99.7 & 92.7 \\
Target chart--condition cells masked & 33.3 & 19.5 & 10.2 \\
\bottomrule
\end{tabular}
\end{table}

On FD, identity weight exceeded 0.5 for 99.89\% of source examples and for 9.32\%, 1.55\%, and 10.99\% of examples from B, A, and D, respectively. Eight of the 24 target chart--condition cells were masked. Proposition~1 certified 889 of the 891 initially correct source predictions. All 889 certified predictions and one additional uncertified prediction remained correct.

Across the three FD target conditions, PACT corrected 889 of 1,042 initially incorrect predictions (85.3\%). Among 1,661 initially correct target predictions, 10 became incorrect (0.60\%). The median identity weight was approximately zero both for target predictions that remained correct and for those that became incorrect. On this stream, the gate distinguished source from target examples, while its identity weight alone provided little indication of whether a target prediction would be damaged.

On HHAR, median source identity weights were 1.00 for preserved correct predictions and 0.06 for predictions that changed from correct to incorrect. On UCI-HAR, the corresponding medians were 0.996 and 0.03. The association between lower identity weight and damaged source predictions can also depend on the initial margin and the magnitude of the applied transport.

\subsection{Sensitivity to gate temperature and chart support}
\label{app:pact-gate-sensitivity}

We vary one routing parameter at a time on FD ($K=8$) and UCI-HAR
($K=4$). The temperature sweep fixes minimum chart support at $m=5$.
The support sweep fixes temperature at $\tau=0.5$.

\begin{table}[h!]
\centering
\caption{Final AMF$_1$ under changes in gate temperature and minimum
chart support. The middle column of each group uses the default
$(\tau,m)=(0.5,5)$.}
\label{tab:pact-routing-sensitivity}
\small
\setlength{\tabcolsep}{6pt}
\renewcommand{\arraystretch}{1.12}
\begin{tabular}{@{}l ccc ccc@{}}
\toprule
& \multicolumn{3}{c}{Temperature ($m=5$)}
& \multicolumn{3}{c}{Minimum support ($\tau=0.5$)} \\
\cmidrule(lr){2-4}\cmidrule(l){5-7}
Dataset & 0.25 & 0.5 & 1.0 & 1 & 5 & 10 \\
\midrule
FD      & 0.9589 & 0.9591 & 0.9599 & 0.9585 & 0.9591 & 0.9517 \\
UCI-HAR & 0.7834 & 0.7911 & 0.8048 & 0.7860 & 0.7911 & 0.7834 \\
\bottomrule
\end{tabular}
\end{table}

Increasing $\tau$ from 0.5 to 1.0 raised AMF$_1$ by 0.08 points on FD
and 1.37 points on UCI-HAR. A minimum support of five achieved the
highest AMF$_1$ among the three thresholds on both datasets.

\subsection{Imbalanced calibration and omitted classes}
\label{app:pact-imbalance}

\rebuttal{We hold the total label budget fixed and vary the class composition of the calibration set. In the imbalance experiment, one class supplies 85\% of the calibration labels. In each omission experiment, one class supplies no target calibration labels. We evaluate held-out recall for the omitted class.}

\begin{table}[h!]
\centering
\caption{Sensitivity to target calibration composition. AMF$_1$ and
recall are on a 0--1 scale. The stream count applies to both datasets.}
\label{tab:pact-imbalance}
\small
\setlength{\tabcolsep}{5pt}
\renewcommand{\arraystretch}{1.12}
\begin{tabular}{@{}llcrr@{}}
\toprule
Calibration setting & Metric & Streams & FD & UCI-HAR \\
\midrule
Class-stratified     & AMF$_1$        & 4 & 0.9575 & 0.7864 \\
85\% from one class  & AMF$_1$        & 4 & 0.8667 & 0.7726 \\
Class 0 omitted      & Class 0 recall & 4 & 0.995  & 0.154  \\
Class 1 omitted      & Class 1 recall & 1 & 0.138  & 0.016  \\
\bottomrule
\end{tabular}
\end{table}

Concentrating 85\% of labels in one class reduced AMF$_1$ by 9.08 points on FD and 1.38 points on UCI-HAR. When class~0 was omitted, its recall was 0.995 on FD and 0.154 on UCI-HAR across four streams. Omitting class~1 yielded low recall on both datasets in the single stream tested. An omitted class can receive a chart-specific correction learned from other labeled classes, but recall depends strongly on the class and dataset.

\section{Supplementary Figures}
\label{sup_figures}

\begin{figure}[h!]
    \centering
    \includegraphics[width=1\linewidth]{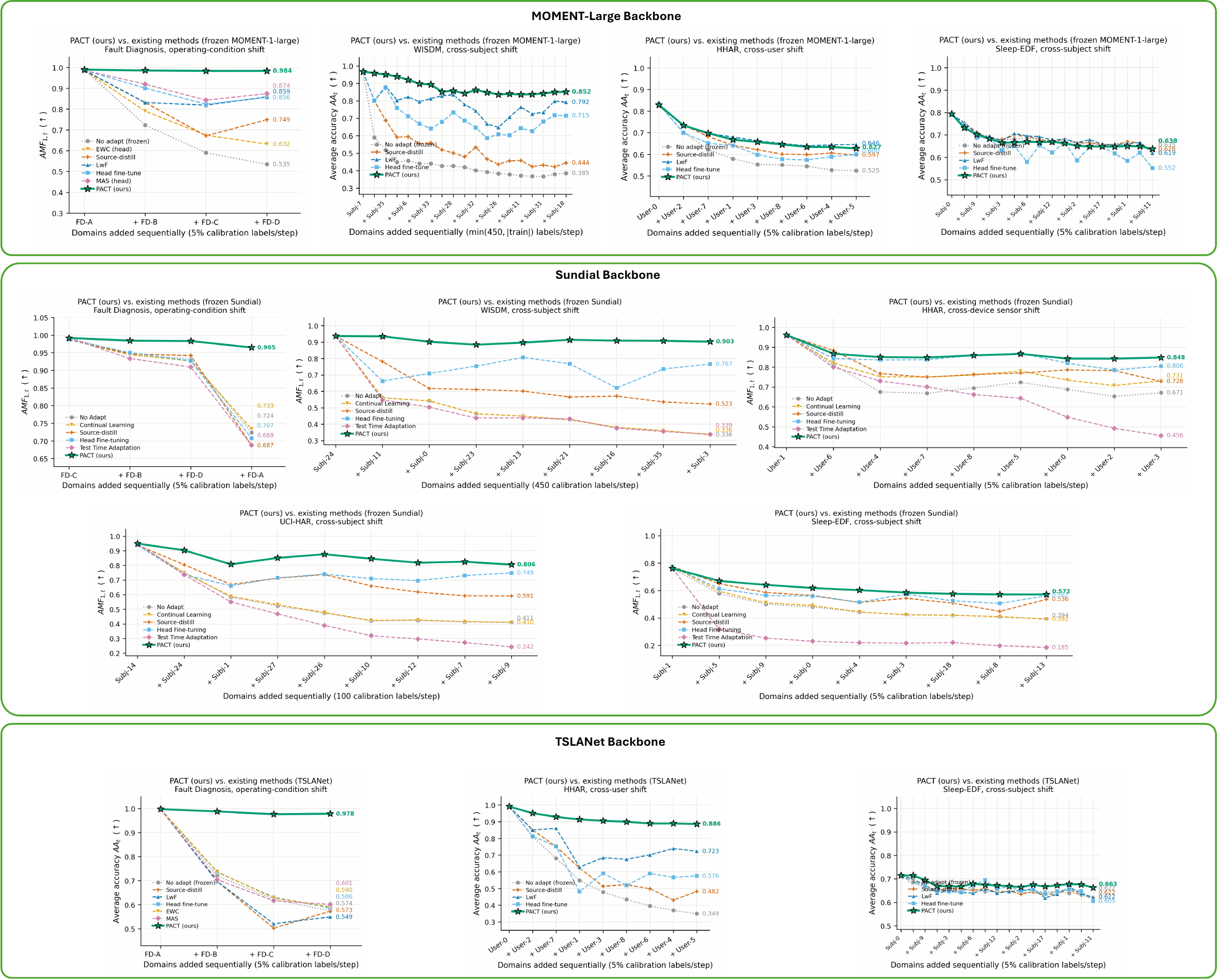}
    \caption{\textbf{Performance throughout domain-incremental streams.} PACT and comparison methods are evaluated with MOMENT-Large, Sundial, and TSLANet along the domain orders shown on the horizontal axes. Each point reports performance over the domains observed so far, using the metric indicated on the vertical axis. Calibration budgets are specified below each panel, and endpoint annotations report final scores for the displayed runs.}
    \label{fig:appendix-learning-curves}
\end{figure}

\begin{figure}[h!]
    \centering
    \includegraphics[width=1\linewidth]{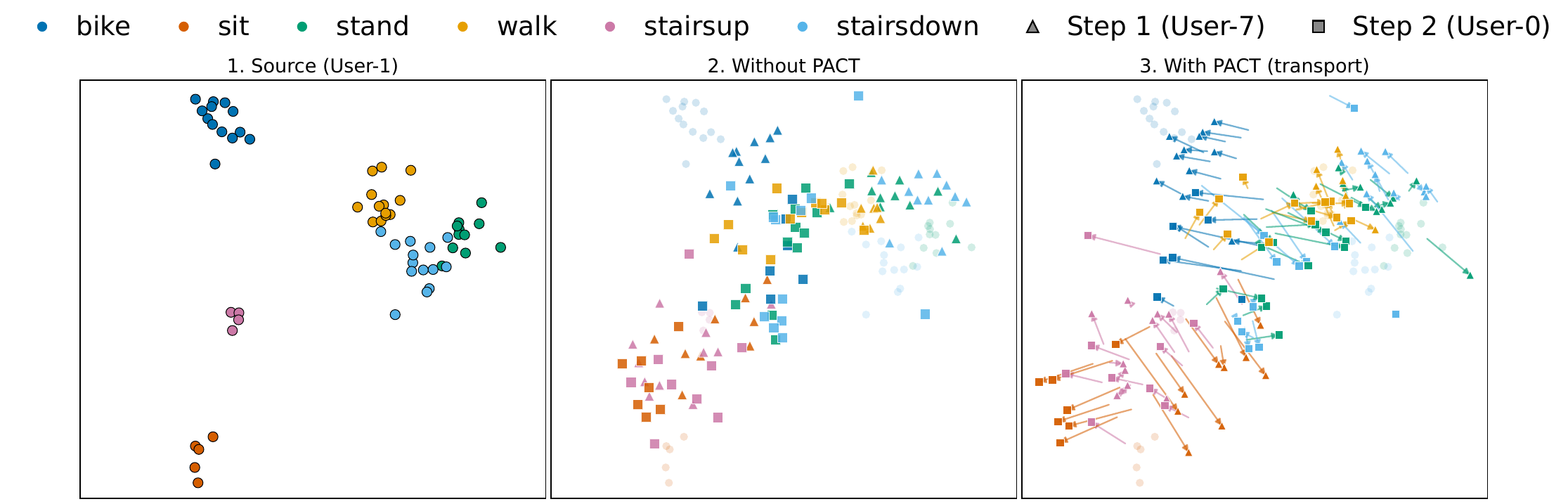}
    \caption{\textbf{Embedding corrections on HHAR.} Left: source embeddings from User-1. Middle: uncorrected embeddings from User-7 and User-0. Right: embeddings after PACT transport, with arrows indicating projected displacements. Colors denote activity classes; triangles and squares identify the two arriving domains. Faded circles provide the source reference in the middle and right panels.}
    \label{fig:appendix-transport}
\end{figure}

\begin{figure}[h!]
    \centering
    \includegraphics[width=1\linewidth]{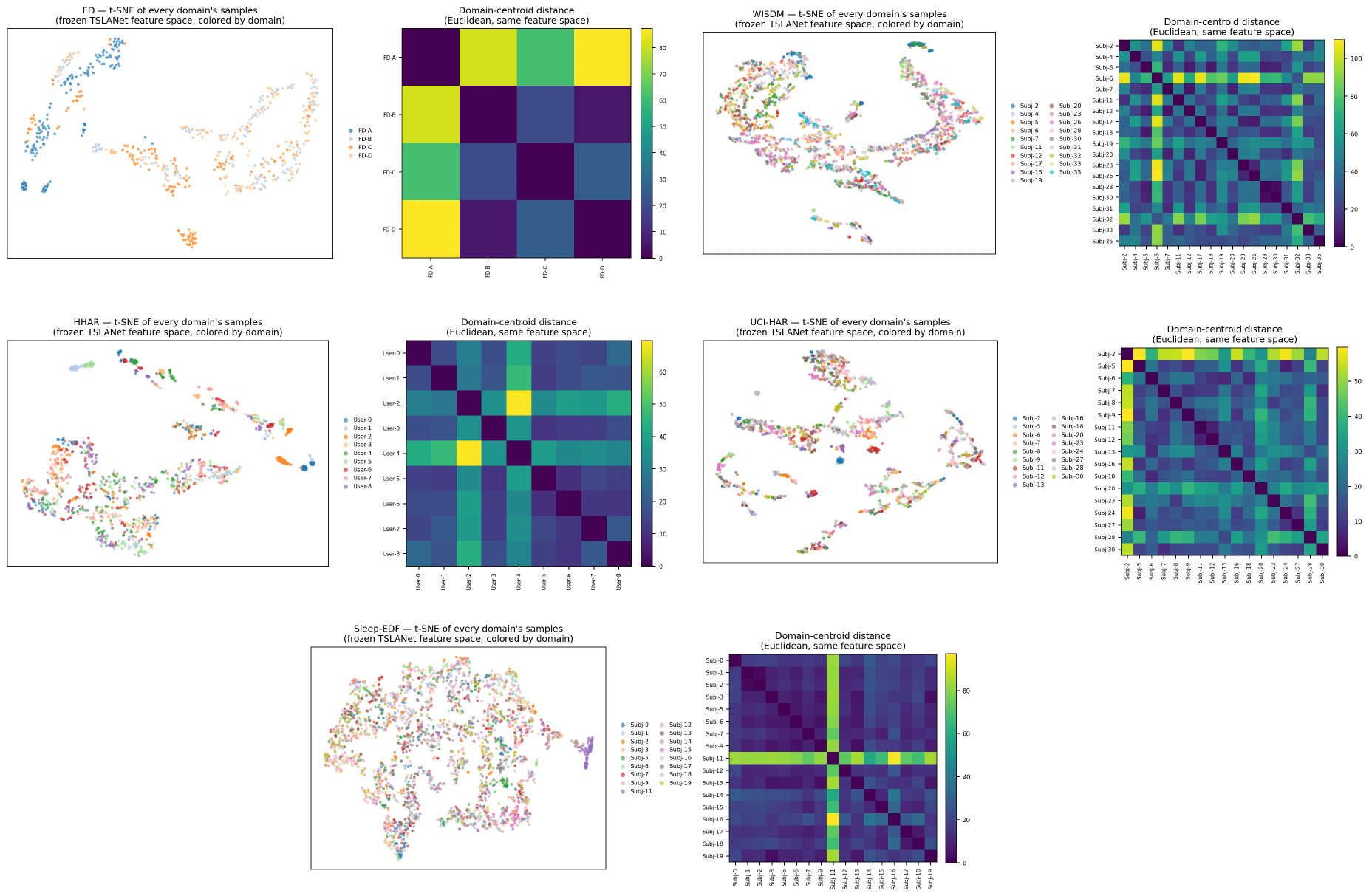}
    \caption{\textbf{Domain structure in frozen representations.} For each dataset, t-SNE projections of frozen TSLANet embeddings are colored by domain, alongside pairwise Euclidean distances between domain centroids in the original embedding space. The projections illustrate varying degrees of domain overlap; centroid distances summarize differences in domain means.}
    \label{fig:appendix-domain-structure}
\end{figure}

\begin{figure}[h!]
    \centering
    \includegraphics[width=1\linewidth]{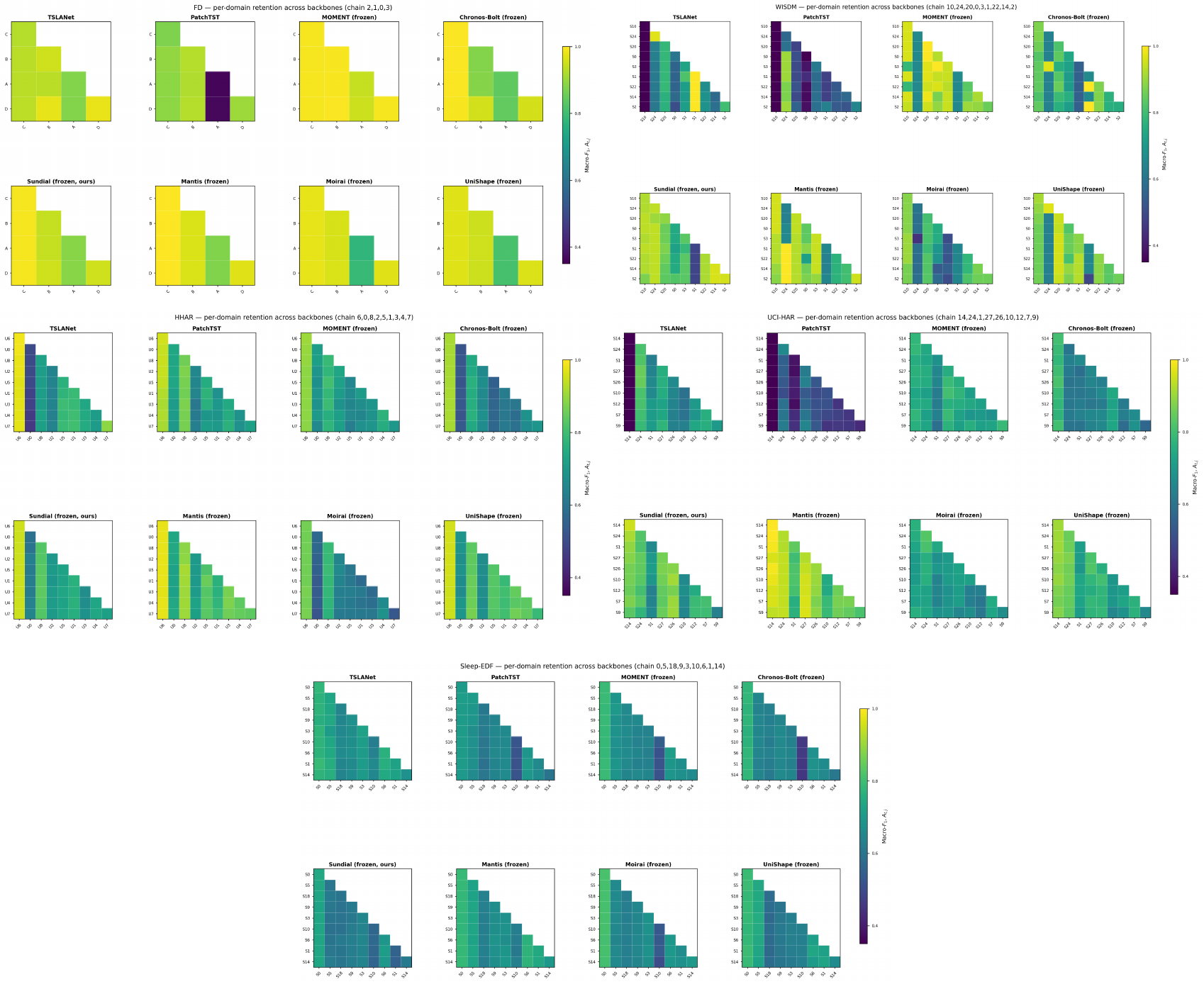}
    \caption{\textbf{Per-domain performance during sequential adaptation.} Matrices show macro-$F_1$ across eight backbones for the domain orders indicated above each dataset. Entry $(t,j)$ reports performance on domain $j$ after learning through step $t$. Diagonal entries record performance when a domain first appears; subsequent entries in each column track its retention. Blank entries correspond to domains not yet observed.}
    \label{fig:appendix-retention}
\end{figure}

We provide additional visualizations of sequential adaptation, embedding corrections, and domain structure. Figures~\ref{fig:appendix-learning-curves} and~\ref{fig:appendix-retention} show performance along individual domain orders, complementing the aggregate results in the main text. Figures~\ref{fig:appendix-transport} and~\ref{fig:appendix-domain-structure} illustrate PACT's corrections and the variation in domain overlap within frozen representations.

\end{document}